%% file: main.tex
\documentclass{aircc}

\usepackage{color}
\usepackage[dvipsnames]{xcolor}

\usepackage[]{graphicx}
\usepackage{tikz}

\usepackage{url}

\usepackage{multirow}
\usepackage[normal, justification=centering]{caption}
\usepackage{diagbox}
\usepackage{makecell}
\usepackage{flexisym}
\usepackage{tablefootnote}
\usepackage{relsize}
\usepackage{subcaption}

\usetikzlibrary{calc,shapes,arrows,decorations.text,positioning,patterns,matrix,chains,decorations.pathreplacing}

\makeatletter
\newcommand{\thickhline}{%
    \noalign {\ifnum 0=`}\fi \hrule height 1pt
    \futurelet \reserved@a \@xhline
}
\newcolumntype{"}{@{\hskip\tabcolsep\vrule width 1pt\hskip\tabcolsep}}
\makeatother

\title
{
Unsupervised robotic sorting: \\ 
Towards autonomous decision making robots
}

\author
{
Joris Gu\'erin, 
St\'ephane Thiery, 
Eric Nyiri and 
Olivier Gibaru
}

\affiliation 
{
Arts et M\'etiers ParisTech, Lille, FRANCE \\
\textit{Correspondance: joris.guerin@ensam.eu}
}

\begin{document}

\maketitle

\begin{abstract}
Autonomous sorting is a crucial task in industrial robotics which can be very challenging depending on the expected amount of automation. Usually, to decide where to sort an object, the system needs to solve either an instance retrieval (known object) or a supervised classification (predefined set of classes) problem. In this paper, we introduce a new decision making module, where the robotic system chooses how to sort the objects in an unsupervised way. We call this problem Unsupervised Robotic Sorting (URS) and propose an implementation on an industrial robotic system, using deep CNN feature extraction and standard clustering algorithms. We carry out extensive experiments on various standard datasets to demonstrate the efficiency of the proposed image clustering pipeline. To evaluate the robustness of our URS implementation, we also introduce a complex real world dataset containing images of objects under various background and lighting conditions. This dataset is used to fine tune the design choices (CNN and clustering algorithm) for URS. Finally, we propose a method combining our pipeline with ensemble clustering to use multiple images of each object. This redundancy of information about the objects is shown to increase the clustering results.
\end{abstract}

\begin{keywords}
Autonomous robotics, Robotic sorting, Image clustering, Ensemble clustering.
\end{keywords}

\input{introduction.tex}
\input{previous_work.tex}
\input{imclust_pipeline.tex}
\input{singleview.tex}

\input{multiview.tex}
\input{conclusion.tex}

\bibliographystyle{IEEEtran}
\bibliography{biblio.bbl}

\end{document}

%% file: introduction.tex
\section{Introduction}
\label{sec:intro}

The problem of automatic sorting has a long history in industry, with the first tomatoes sorting system dating back to the 70's \cite{first_sorting_system}. Since then, it has received a lot of attention, with important focus on combining computer vision and robotics manipulator to solve the pick-and-place task \cite{survey_automatic_sorting}. 

Although it was among the first robotics tasks, designing a pick-and-place application is challenging and requires solving multiple subtasks. If the objects to be sorted are cluttered, the system first needs to segment the scene and identify the different objects \cite{object_segmentation}. Then, the systems must use various sensors (2D or 3D cameras, color sensors, bar code readers, ...) to gather data about the different objects. These data are used to find a grasping strategy for the object \cite{grasping_survey} and to decide where to place the object. Naturally, all along this process, smart motion planning and control is also required \cite{handbook_robotics}.

Although every brick of the pick-and-place pipeline are interesting and challenging problems, the scope of this paper is limited to the decision making subproblem, i.e. how to decide where to place an object after grasping it? As discussed later in Section 2, previous implementations of automatic sorting deal with objects which are either known or belong to some predefined class. In machine learning terminology, one could say that the robot intelligence usually resides in solving either an instance retrieval or a supervised classification problem. In this paper, a new kind of pick-and-place application is introduced, consisting in sorting unknown objects, which do not belong to any predefined class. Given a set of previously unseen objects, the robotic system needs to sort them such that objects stored together are similar and different from other objects. Such problem definition is the definition of a clustering problem in which the outputs consist in physically sorted objects, hence, we call this problem Unsupervised Robotic Sorting (URS).

\begin{figure}[t]
    \centering
    
    \begin{tikzpicture}
    \node[inner sep=0pt] (im0) at (0,0)
        {\includegraphics[width=.3\textwidth]{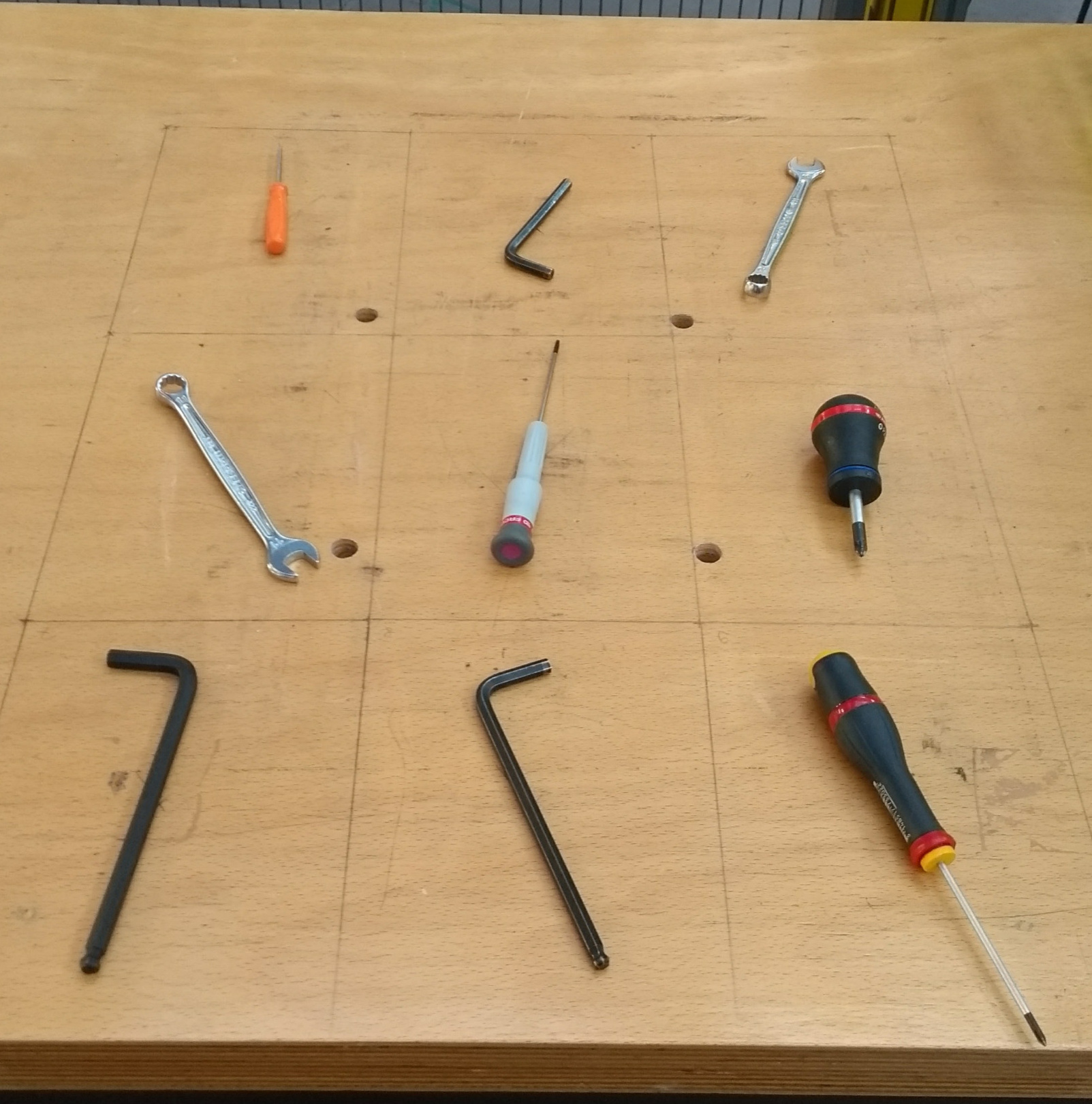}};
    \node[inner sep=0pt] (out) at (8, 0)
        {\includegraphics[width=.4\textwidth]{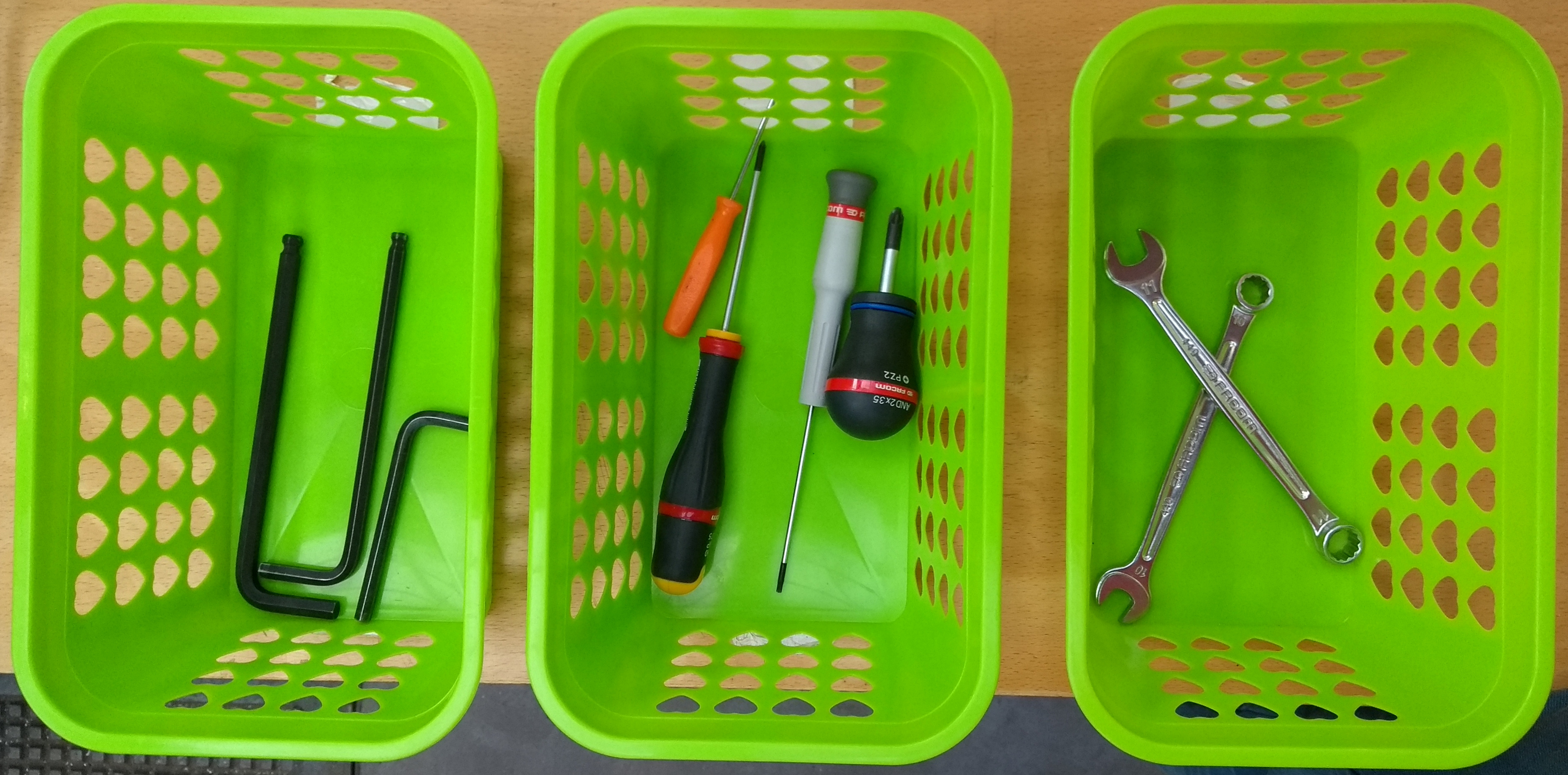}};
    \draw[->,line width=0.5mm] (im0.east) -- (out.west)
        node[above, midway] {Unsupervised} node[below, midway] {Robot Sorting};
    \end{tikzpicture}
    
    \caption{An instance of the unsupervised robotic sorting problem.}
    \label{fig:intro_example}
\end{figure}

The above statement of the URS problem neither defines the notion of similarity nor the number of groups to be made. Different definitions of these concepts defines different instances of the URS problem. In this paper, we propose a solution to solve a URS in which the number of groups is imposed by the number of storage spaces available and the notion of similarity is defined by ``how a human would solve the problem''. Figure \ref{fig:intro_example} shows a practical example of inputs and expected output. In practice, each object is sensed with a camera and represented as an image, the semantic content of the image is extracted using a pretrained convolutional neural network (CNN) (\cite{feature_extraction}) and the classification is done using a ``standard'' clustering algorithm. A complete description of the algorithm used is given in Section 3. A video of our implementation can be found at \url{https://youtu.be/NpZIwY3H-gE}.

URS has various potential applications in industry such as organizing workspaces before the work day and sorting unsorted goods before packaging. However, to be useful in a real industrial shop floor, a URS implementation needs to work under a broad range of conditions. Object poses, background and lighting conditions are factors which vary a lot among industrial settings and they can impact widely the quality of a computer vision application. Hence, we introduce a new image dataset specifically designed to evaluate industrial robustness of our application. The dataset contains object which can be found on a shop-floor, under different orientations, background and lighting conditions. Together with artificial lighting modification, the dataset is used to choose the most robust components of the proposed clustering pipeline (CNN architecture and clustering algorithm). 

It is also important to note that URS consists in sorting real objects, not only images. Hence, we propose to further improve the robustness of our implementation by introducing redundancy of information for each objects. After choosing the final elements of the image clustering pipeline, we propose to embed it in a multi-view clustering pipeline, taking as input several images, under different orientations, of each object. Leveraging ensemble clustering, we show that URS robustness can be increased using multiple views of each objects.

The main contributions of this paper consist in the formalization, together with a working implementation, of the unsupervised robotic sorting problem. Such implementation is, to the best of our knowledge, the first application of image-set clustering to robotics and constitutes a new step towards autonomous decision-making robots. To validate our solution, a new image dataset is proposed. Such dataset, which appears to be challenging for the image-set clustering task, is made publicly available in order to help other researchers to test the robustness of their image-set clustering algorithms. By implementing a multi-view clustering algorithm based on ensemble clustering, we also show that URS robustness can be improved by using multiple images of the objects.

The paper is organized as follows. In Section 2, we present a literature review on image-set clustering, multi-view clustering and decision making in robotic sorting. In Section 3, we define in details the vision pipeline proposed to solve the URS problem and validate our choice on several public datasets. In Section 4, we describe the complete URS implementation, introduce the validation dataset and test our approach on complex cases. Then we propose, in Section 5, a multi-view approach, based on ensemble clustering, to improve the robustness of URS. Finally, Section 6 discusses our findings and future working directions.

%% file: previous_work.tex
\section{Previous Work}
\label{sec:prev_work}

\subsection{Decision making in autonomous sorting}
    \label{sec:prev_decision_making}

Although scene segmentation, robotic grasping, motion planning and control are crucial problems, which have received major attention recently, the scope of this paper is limited to the decision making subproblem of the pick-and-place pipeline. We classify the approaches to decision making into three categories: known object (instance retrieval), unknown object member of a known class (supervised classification), unknown object and no predefined class (unsupervised classification). Different approaches in solving such decision-making problems differ in the kind of sensing devices used, as well as the algorithmic choices.

\textbf{Instance retrieval.} Several implementations have considered the problem of recognizing each object as a member of a known database, and sorting it according to an associated predefined rule. In \cite{rfid}, RFID markers are used to recognize and sort objects. An implementation of a surgical tools sorting application using barcode reading and template matching to find and sort objects from a cluttered scene is proposed in \cite{surgical_tools_sorting}. Finally, the application presented in \cite{realworld_templatematching} sorts real-world objects (usb, glue-stick, ...) with a serial manipulator using template matching.

\textbf{Supervised classification.} Many robotics supervised classification applications sort objects based on simple rules (thresholding) applied to simple features. For example, \cite{cube_color, color_sensor, color_sorting_scara} all use color features extracted from either images or color sensors, \cite{color_shape} extend these approaches by adding shape features. In \cite{clutered_duplo_colorshape}, robot manipulation is used to unclutter Duplo bricks and sorts them according to their length and color. In \cite{fault_sorting}, a classical computer vision framework is implemented to carry out faulty parts removal. A robotic system able to recycle construction wastes using features provided by metal detectors and sensors sensitive to visual wavelengths is presented in \cite{construction_recycling}. Recently, researchers have started to use CNN methods to carry out supervised classification in robotic sorting. The authors of \cite{rgbd_objectrec} have used a CNN on RGBD images to find the class of an object and sort it. In \cite{fastrcnn_sorting}, a fast RCNN proposal network is combined with a pretrained VGG16 CNN architecture to jointly localize and sort objects with a robot manipulator.

\textbf{Unsupervised classification.} In previous work, we proposed a first implementation of unsupervised robotic sorting \cite{lego_clustering}. The approach consisted in clustering a set of Lego bricks using color and length features extracted with a computer vision pipeline. We also note that this paper is an extension of our previous work on transferring pretrained CNN features to unsupervised tasks \cite{aifu}, where the single-view implementation of URS was discussed.

\subsection{Image-set clustering} 

Given a set of unlabeled images, the image-set clustering problem consists in finding subsets of images based on their content: two images representing the same object should be clustered together and separated from images representing other objects (Figure \ref{fig:intro_example}). This problem should not be confused with image segmentation \cite{segmentation}, which is also sometimes called image clustering. Image-set clustering has been used to solve various problems over the last two decades. Web-scale image clustering has been tackled with variants of K-Means in both \cite{webscale1} and \cite{webscale2}. Concept discovery in image categories is solved using spectral clustering in \cite{conceptdiscovery}. In \cite{storyline} a clustering based solution to story-line reconstruction in streams of images is proposed and in \cite{alternating_opt_clust1} image-set clustering is applied to large-scale medical image annotation.

The first successful methods focused on feature selection and used sophisticated algorithms to deal with complex features. For instance, in \cite{GMM}, images are represented by Gaussian Mixture Models fitted to the pixels and the set of images is clustered using the Information Bottleneck (IB) method. Features resulting from image joint segmentation and sequential IB for clustering are used in \cite{segmentation_image_set}. In \cite{commonality_clustering}, commonality measures are defined from Bags of Features with local representations (SIFT, SURF, ...) and used for agglomerative clustering. Recently, image-set clustering algorithms have started to use features extracted with deep CNN. \cite{webscale2} applies a variant of approximate K-means on features extracted with a pretrained AlexNet and \cite{infinite_ensemble} uses deep auto-encoders combined with ensemble clustering to generate feature representations suitable for clustering. As of today, a new family of methods consisting in learning jointly the clusters and the representation using alternating optimization has established a new baseline on the performance of image-set clustering algorithms (\cite{jule, alternating_opt_clust1, alternating_opt_clust2}).

Literature review of image-set clustering algorithms raises two useful remarks for the URS implementation. First, objects representations should be extracted from images using a pretrained CNN. Then, given the scale of the clustering problems to solve $(\#images<100)$, it is unlikely that jointly optimizing the CNN parameters would improve the final sorting results. We give a detailed description of our image-set clustering pipeline as well as experimentally motivated design choices in the next section.

\subsection{Multi-view clustering}
    
Multi-view learning improves results classification results by leveraging complementary information from multiple views of a dataset, which can be obtained by various sensors or represented with different descriptors. Recently, multi-view clustering (MVC) has received a lot of attention. In \cite{multiview_concat} different loss functions are introduced and applied on the concatenated views. In both \cite{multiview_latent1} and \cite{multiview_latent2} lower dimensional subspaces are learned before clustering with standard methods. Recently, it has been shown in \cite{multiview_ensemble} that good results can be obtained by embedding MVC in the ensemble clustering framework \cite{ensemble_survey}. The authors leveraged the different views to generate independent partitions and then used a co-association based method to obtain the consensus.

%% file: imclust_pipeline.tex
\section{Image clustering pipeline}
\label{sec:pipeline}

In this section, we discuss the vision module of our URS application. First, the generic form of the image-set clustering pipeline is described. Then, we compare several sets of hyperparameters on the Pascal VOC 2007 test set. These preliminary results are used to prune the bad combinations and select only relevant design choices for further testings in section 4. Then, we validate the relevance of the chosen approach by comparing the best pipeline obtained with VOC against other approaches from the literature.

\subsection{Pipeline description}

The pipeline we propose for image set clustering is fairly straightforward. It consists in extracting deep features from all the images in the set, by using a deep convolutional neural network pretrained on a large supervised dataset for image classification and then apply a ``standard" clustering algorithm to these features. The proposed vision module can be visualized later in Figure \ref{subfig:vismod}.

To implement this unsupervised image classification pipeline, we first need to answer four questions:
\begin{itemize}
    \item What dataset should be used for pretraining?
    \item What CNN architecture should we use?
    \item Which layer output should be chosen for feature extraction?
    \item What clustering algorithm should be used?
\end{itemize}

As of today, ImageNet \cite{imagenet} is the only very large labelled public dataset which has enough variety in its classes to be a good feature extractor for a variety of tasks. It has been shown than features extracted from CNN pretrained on ImageNet can be successfully transferred to a variety of tasks \cite{feature_extraction}. In addition, there are plenty of CNN pretrained on ImageNet already available online. Hence, we naturally use CNNs pretrained on ImageNet. The three other questions are answered experimentally. We use the VOC2007 \cite{voc} test set without labels to compare performances of the different options and select a small subset of parameters for further investigations on URS.

To ease development and accelerate implementation, we compare the Keras \cite{keras} implementations of ResNet50 \cite{resnet}, InceptionV3 \cite{inception}, VGG16, VGG19 \cite{vgg} and Xception \cite{xception} with the pretrained weights provided by Keras. For the clustering algorithms, we use the scikit-learn \cite{sklearn} implementations of K-means (KM) \cite{kmeans++}, Minibatch K-means (MBKM) \cite{mbkm}, Affinity Propagation (AP) \cite{affinity}, Mean Shift (MS) \cite{mean_shift}, Agglomerative Hierarchical Clustering (AC) \cite{agglomerative} and DBScan (DBS) \cite{dbscan}. For each CNN, the layers after which the features are extracted can be found in Table \ref{tab:voc} (Layers names are the same as in the Keras implementations).

In the image-set clustering problem, the expected classes are represented by objects present on the picture and for this reason we need semantic information, which is present in the final layers of the network. Thus, we only choose layers among the last layers of the networks. On the one hand, the last one or two layers might provide better results as their goal is to separate the data (at least for the fully-connected layers). On the other hand, the opposite intuition is also relevant as we can imagine that these layers are too specialized to be transferable. These two contradictory arguments motivated the following experiment.

We also note that the test set of VOC2007 has been modified for this validation. We removed all the images containing two or more labels in order to have ground truth to compute Normalized Mutual Information (NMI) scores. Indeed, if an image has several labels we cannot judge if the clustering pipeline prediction is correct or not. We note VOC2007-SL (single label) the modified VOC2007 test set.

\subsection{Hyperparameters choice}

To answer the questions stated above, we try to cluster the VOC2007-SL set using all combinations of CNN architectures, layer choices and clustering algorithms. To compare performances, we use NMI scores, a standard clustering evaluation metric. We also report clustering time for completeness. Only scikit-learn default hyperparameters of the different clustering algorithms are used, which illustrate the simplicity of this approach. For KM and MBKM, as the results depend on a random initialization, experiments are run ten times and the results reported are averaged over the different runs.

\input{table_voc.tex}

These experiments show that, for every CNN architecture and clustering algorithm pairs, the results get better when we choose layers which are further in the architecture. Hence, from now on, we only consider the last layer before classification for each network. For example, in the rest of the paper, VGG16 now denotes the features extracted from fc2 in the VGG 16 architecture. Looking at the results, we also keep only two clustering algorithm (KM and AC) for our URS testings. This leaves us with 10 configurations to test in section 4.

In the next section, the relevance of the proposed pipeline is further evaluated. We compare the most successful pipeline (Agglomerative Clustering on features extracted from the final layer of an Xception CNN) against other methods from the literature on other public datasets.

\subsection{Literature comparison}

To justify the use of the proposed clustering routine, we compare its results against other image clustering methods on several public datasets, which are described in Table \ref{tab:pub_data}. The tasks involved by these datasets are different from each others (Face recognition, grouping different objects, recognizing different pictures of the same object). In addition, the content of the classes differs from the ones in ImageNet. For these reasons, the four datasets constitute a good benchmark to quantify the robustness of transfer learning for unsupervised object categorization.

\input{table_datasets.tex}

We propose a comparison with the results reported in the following papers:
\begin{itemize}
    \item \cite{commonality_clustering} proposes different clustering algorithms applied on bags of features. In Table \ref{tab:pub_res}, we note "BoF" the best results obtained by such pipeline on the different datasets.
    
    \item \cite{infinite_ensemble} proposes a method called infinite ensemble clustering (IEC). In the paper, IEC algorithm is compared to several other deep clustering algorithms and ensemble clustering algorithms. In Table \ref{tab:pub_res}, we report the best results obtained using Deep Clustering (DC) and Ensemble Clustering (EC) for each datasets. We note that for VOC2007-5-ML, \cite{infinite_ensemble} also uses deep features as clustering inputs (the CNN used is not reported).
    
    \item \cite{jule} proposes a method called Joint Unsupervised Learning (JULE) of Deep Representations and Image Clusters, based on Alternating optimization between clustering and weight optimization of the CNN feature extractor. Results from this work are reported in Table \ref{tab:pub_res}.
\end{itemize}

For each dataset groundtruth is known as they are intended for supervised classification. We compute both NMI scores and purity for each dataset/method pair.

\input{table_pipeline_eval.tex}

Table \ref{tab:pub_res} shows that features extracted by the final layer of Xception combined with Agglomerative Clustering performs better than or close to state-of-the-art methods at unsupervised object classification as well as fine-grained image clustering (ORL). Results on the ORL dataset are interesting as they show that pretrained Xception is able to classify different faces without supervision, although ImageNet does not deal with human faces at all.

We underline the very good results of JULE \cite{jule} at clustering COIL100. It appears that fine tuning feature extraction using alternating optimization is a good way of improving clustering results. However, the simple approach proposed here, which provides slightly lower results, still keeps the advantage of being very fast (as it only requires to evaluate the network once for each sample and apply AC), which is useful for our application. Moreover, the very small sample size of URS problems makes retraining impractical. From our attempts of using JULE for URS, it seems that the approach in this paper is more relevant.

It is also interesting to notice that \cite{infinite_ensemble} is also using CNN features for clustering VOC2007-5-ML. The fact that our pipeline outperform largely DC and EC for this dataset illustrate that when using transfer learning for a new problem, careful selection of the CNN used is important, which is often neglected.

%% file: table_voc.tex
\begin{table}[!ht]
\caption{NMI scores (in black) and time in seconds (in blue, italics) on Pascal VOC2007-SL test set using different CNN, different output layers and different clustering algorithms. (Layers names are the same as in the Keras implementations).}
\label{tab:voc}
\centering
\vspace{\baselineskip}
    \scalebox{0.9}{
    \begin{tabular}{cc"c|c|c|c|c|c}
        
        \multicolumn{2}{c"}{} & KM & MBKM & AP & MS & AC & DBS\tablefootnote{DB-Scan poor results might come from the default parameters and using different configurations might improve clustering, but this is out of the scope of this paper.}\tabularnewline
        \Xhline{2\arrayrulewidth}
        \multirow{6}{*}{Inception V3} & \multirow{2}{*}{mixed9} & 0.108 & 0.105 & 0.219 & 0.153 & 0.110 & 0 \tabularnewline 
        & & \textit{\textcolor{Blue}{374}} & \textit{\textcolor{Blue}{7.3}} & \textit{\textcolor{Blue}{12.7}} & \textit{\textcolor{Blue}{16281}} & \textit{\textcolor{Blue}{525}} & \textit{\textcolor{Blue}{138}} \tabularnewline \cline{2-8}
        
        & \multirow{2}{*}{mixed10} & 0.468 & 0.401 & 0.442 & 0.039 & 0.595 & 0 \tabularnewline 
        & & \textit{\textcolor{Blue}{609}} & \textit{\textcolor{Blue}{5.1}} & \textit{\textcolor{Blue}{8.5}} & \textit{\textcolor{Blue}{12126}} & \textit{\textcolor{Blue}{525}} & \textit{\textcolor{Blue}{119}} \tabularnewline \cline{2-8}
        
        & \multirow{2}{*}{avg\_pool} & 0.674 & 0.661 & 0.621 & 0.024 & 0.686 & 0 \tabularnewline 
        & & \textit{\textcolor{Blue}{6.3}} & \textit{\textcolor{Blue}{0.2}} & \textit{\textcolor{Blue}{7.7}} & \textit{\textcolor{Blue}{230}} & \textit{\textcolor{Blue}{8.5}} & \textit{\textcolor{Blue}{1.8}} \tabularnewline \hline
        
        \multirow{2}{*}{Resnet 50} & \multirow{2}{*}{avg\_pool} & 0.6748 & 0.641 & 0.587 & 0.043 & 0.640 & 0 \tabularnewline 
        & & \textit{\textcolor{Blue}{7.0}} & \textit{\textcolor{Blue}{0.1}} & \textit{\textcolor{Blue}{4.6}} & \textit{\textcolor{Blue}{197}} & \textit{\textcolor{Blue}{8.0}} & \textit{\textcolor{Blue}{1.9}} \tabularnewline \hline
        
        \multirow{8}{*}{VGG 16} & \multirow{2}{*}{block4\_pool} & 0.218 & 0.085 & 0.133 & 0.124 & 0.277 & 0 \tabularnewline 
        & & \textit{\textcolor{Blue}{278}} & \textit{\textcolor{Blue}{3.6}} & \textit{\textcolor{Blue}{6.0}} & \textit{\textcolor{Blue}{10010}} & \textit{\textcolor{Blue}{391}} & \textit{\textcolor{Blue}{82.8}} \tabularnewline \cline{2-8}
        
        & \multirow{2}{*}{block5\_pool} & 0.488 & 0.048 & 0.262 & 0.194 & 0.530 & 0 \tabularnewline 
        & & \textit{\textcolor{Blue}{78}} & \textit{\textcolor{Blue}{1.1}} & \textit{\textcolor{Blue}{9.3}} & \textit{\textcolor{Blue}{2325}} & \textit{\textcolor{Blue}{99}} & \textit{\textcolor{Blue}{21}} \tabularnewline \cline{2-8}
        
        & \multirow{2}{*}{fc1} & 0.606 & 0.458 & 0.421 & 0.187 & 0.617 & 0 \tabularnewline 
        & & \textit{\textcolor{Blue}{17}} & \textit{\textcolor{Blue}{0.2}} & \textit{\textcolor{Blue}{4.8}} & \textit{\textcolor{Blue}{365}} & \textit{\textcolor{Blue}{17}} & \textit{\textcolor{Blue}{3.8}} \tabularnewline \cline{2-8}
        
        & \multirow{2}{*}{fc2} & 0.661 & 0.611 & 0.551 & 0.085 & 0.673 & 0 \tabularnewline 
        & & \textit{\textcolor{Blue}{16}} & \textit{\textcolor{Blue}{0.2}} & \textit{\textcolor{Blue}{4.3}} & \textit{\textcolor{Blue}{373}} & \textit{\textcolor{Blue}{15.9}} & \textit{\textcolor{Blue}{3.8}} \tabularnewline \hline
        
        \multirow{8}{*}{VGG 19} & \multirow{2}{*}{block4\_pool} & 0.203 & 0.139 & 0.124 & 0.135 & 0.234 & 0 \tabularnewline 
        & & \textit{\textcolor{Blue}{220}} & \textit{\textcolor{Blue}{3.7}} & \textit{\textcolor{Blue}{6.3}} & \textit{\textcolor{Blue}{10298}} & \textit{\textcolor{Blue}{388}} & \textit{\textcolor{Blue}{83}} \tabularnewline \cline{2-8}
        
        & \multirow{2}{*}{block5\_pool} & 0.522 & 0.321 & 0.250 & 0.198 & 0.540 & 0 \tabularnewline 
        & & \textit{\textcolor{Blue}{74}} & \textit{\textcolor{Blue}{0.9}} & \textit{\textcolor{Blue}{9.3}} & \textit{\textcolor{Blue}{2353}} & \textit{\textcolor{Blue}{97}} & \textit{\textcolor{Blue}{20}} \tabularnewline \cline{2-8}
        
        & \multirow{2}{*}{fc1} & 0.607 & 0.471 & 0.449 & 0.188 & 0.628 & 0 \tabularnewline 
        & & \textit{\textcolor{Blue}{17}} & \textit{\textcolor{Blue}{0.2}} & \textit{\textcolor{Blue}{9.3}} & \textit{\textcolor{Blue}{365}} & \textit{\textcolor{Blue}{17}} & \textit{\textcolor{Blue}{3.9}} \tabularnewline \cline{2-8}
        
        & \multirow{2}{*}{fc2} & 0.672 & 0.615 & 0.557 & 0.083 & 0.671 & 0 \tabularnewline 
        & & \textit{\textcolor{Blue}{15}} & \textit{\textcolor{Blue}{0.2}} & \textit{\textcolor{Blue}{5.6}} & \textit{\textcolor{Blue}{391}} & \textit{\textcolor{Blue}{17}} & \textit{\textcolor{Blue}{3.9}} \tabularnewline \hline
        
        \multirow{6}{*}{Xception} & \multirow{2}{*}{block13\_pool} & 0.376 & 0.264 & 0.351 & 0.044 & 0.473 & 0 \tabularnewline 
        & & \textit{\textcolor{Blue}{410}} & \textit{\textcolor{Blue}{4.8}} & \textit{\textcolor{Blue}{10}} & \textit{\textcolor{Blue}{9677}} & \textit{\textcolor{Blue}{403}} & \textit{\textcolor{Blue}{87}} \tabularnewline \cline{2-8}
        
        & \multirow{2}{*}{block14\_act} & 0.574 & 0.428 & 0.584 & 0.071 & 0.634 & 0 \tabularnewline 
        & & \textit{\textcolor{Blue}{935}} & \textit{\textcolor{Blue}{10}} & \textit{\textcolor{Blue}{10}} & \textit{\textcolor{Blue}{24809}} & \textit{\textcolor{Blue}{820}} & \textit{\textcolor{Blue}{180}} \tabularnewline \cline{2-8}
        
        & \multirow{2}{*}{avg\_pool} & 0.692 & 0.636 & 0.636 & 0.052 & \textbf{0.726} & 0 \tabularnewline 
        & & \textit{\textcolor{Blue}{7.1}} & \textit{\textcolor{Blue}{0.1}} & \textit{\textcolor{Blue}{4.9}} & \textit{\textcolor{Blue}{201}} & \textit{\textcolor{Blue}{\textbf{8.5}}} & \textit{\textcolor{Blue}{5.5}} \tabularnewline
        
    \end{tabular}}
    
   
    

   
\end{table}

%% file: table_datasets.tex
\begin{table}[!ht]
\caption{Several key features about the datasets used for the vision pipeline validation.} 
\label{tab:pub_data}
\centering

    \begin{tabular}{c|c|c|c|c}
        & Problem type & Image Size & \# Classes & \# Instances\tabularnewline \hline
        COIL100 \cite{coil100} & Object recognition & $128 \times 128$ & 100 & 7201 \tabularnewline \hline
        Nisters \cite{nister} & Object recognition & $640 \times 480$ & 2550 & 10200 \tabularnewline \hline
        ORL \cite{orl} & Face recognition & $92 \times 112$ & 40 & 400 \tabularnewline \hline
        VOC2007-5-ML \tablefootnote{The data used for VOC2007 in \cite{infinite_ensemble} are irrelevant for clustering with hard assignment. The VOC2007 subset used in \cite{infinite_ensemble} contains images with several objects but only one label. However, we still ran our clustering method on this data to be able to compare results. This second modified VOC2007 set is denoted VOC2007-5-ML (5 classes - multiple labels)} & Object recognition & variable & 5 & 3376
    \end{tabular}

\vspace*{3pt}
   
\end{table}

%% file: table_pipeline_eval.tex
\begin{table}[!ht]
\caption{NMI scores and purity comparison on various public datasets. (\footnotesize{A result that is not reported in the papers cited above is denoted N.R.})} 
\label{tab:pub_res}
\centering

    \begin{tabular}{c|c|c|c|c|c}
        \multicolumn{6}{c}{NMI scores}\vspace{1mm}\tabularnewline
        & DC & EC & BoF & JULE & Xception + AC\tabularnewline \hline
        COIL100 & 0.779 & 0.787 & N.R. & 0.985 & 0.951 \tabularnewline \hline
        Nisters & N.R. & N.R. & 0.918 & N.R. & 0.991 \tabularnewline \hline
        ORL & 0.777 & 0.805 & 0.878 & N.R. & 0.93 \tabularnewline \hline
        VOC2007-5-ML & 0.265 & 0.272 & N.R. & N.R. & 0.367
    \end{tabular}
    
    \vspace*{10pt}
    
    \begin{tabular}{c|c|c|c}
        \multicolumn{4}{c}{Purity}\vspace{1mm}\tabularnewline
        & DC & EC & Xception + AC\tabularnewline \hline
        COIL100 & 0.535 & 0.546 & 0.882 \tabularnewline \hline
        Nisters & N.R. & N.R. & 0.988 \tabularnewline \hline
        ORL & 0.578 & 0.630 & 0.875 \tabularnewline \hline
        VOC2007-5-ML & 0.513 & 0.536 & 0.622
    \end{tabular}

\vspace*{3pt}
   
\end{table}

%% file: singleview.tex
\section{URS implementation}
\label{sec:singleview}

\subsection{Physical implementation}

\begin{figure}[!ht]
    \centering
    
    \begin{subfigure}{\textwidth}
    
    \input{full_pipeline.tex}
    \vspace{\abovecaptionskip}
    \caption{Unsupervised robotic sorting pipeline.}
    
    \end{subfigure}

    \vspace{1.5\floatsep}
    
    \begin{subfigure}{\textwidth}
    
    \input{vision_unit.tex}
    \vspace{\abovecaptionskip}
    \caption{Vision module description.}
    \label{subfig:vismod}
    
    \end{subfigure}
    
    \caption{Unsupervised robot sorting implementation.}
    \label{fig:implementation}
\end{figure}
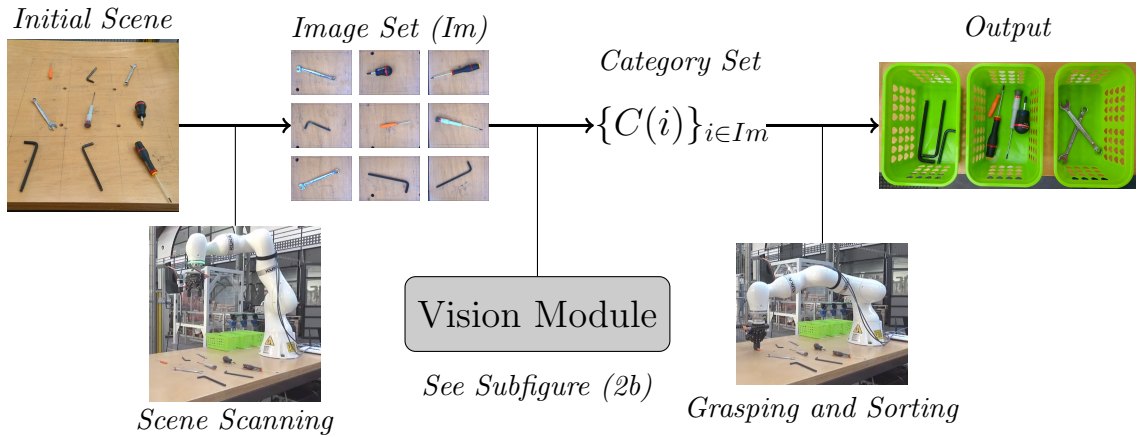
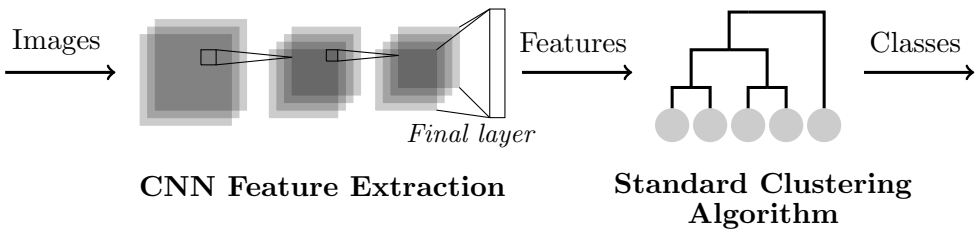

In our implementation of URS, we use a KUKA LBR iiwa robot. Both a camera and a parallel gripper are mounted on the end-effector. First, different previously unseen objects are placed in the robot workspace. The robot scans the scene and gathers pictures of each object to be sorted. The above computer vision pipeline is applied to the image set and returns a bin number for each object. Finally, the robot physically sorts the objects according to the clustering results. Figure \ref{fig:implementation} shows a schematic view of the system, and a video of the implementation can be viewed at \url{https://youtu.be/NpZIwY3H-gE}.

The application was implemented in a shopfloor with unmastered lighting (glass roof) and a wooden table background. We tested the URS using sets of different tools (screw drivers, allen keys, flat keys, clamps, ...) and the goal was to correctly group together the different models of the same tool. For this particular task, we recorded only one misclassification in over 100 runs, which was due to very high brightness (white image).

\subsection{Robustness evaluation and fine tuning of the clustering pipeline}
\subsubsection{Dataset}

To test the robustness of the URS application, we build a challenging dataset for image clustering, composed of pictures of objects which can be found in industrial workstations. Objects are chosen from seven classes, and pictures of each objects are taken under five different background/lighting conditions (BLC). This dataset appears to be challenging for image-set clustering because of the BLC variations but also because some classes have low intra-cluster similarity (usb) and extra-cluster similarity between some classes is relatively high (pens/screws). For each object and each BLC, the dataset contains four pictures under different position/orientation. This is used to test the robustness to objects pose variation and to test the multi-view method in Section 5.

The dataset statistics are summarized in Table \ref{tab:new_datastat} and sample images, illustrating the different objects and BLC, can be seen on Figure \ref{fig:dataset}. The dataset, together with its description, can be downloaded at \url{https://github.com/jorisguerin/toolClustering_dataset}.

\input{table_dataset.tex}
\input{fig_imageSample.tex}

\begin{figure}[t]
    \centering
    
    \begin{tikzpicture}
    \node[inner sep=0pt] (im0) at (0,0)
        {\includegraphics[width=.17\textwidth]{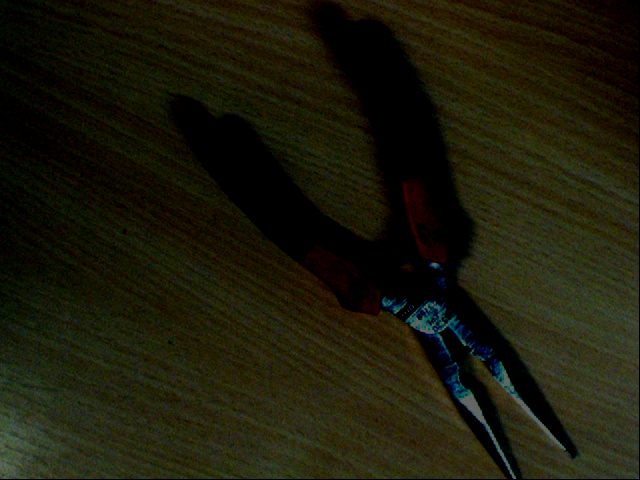}};
    \node[inner sep=0pt] (im0) at (.2\textwidth,0)
        {\includegraphics[width=.17\textwidth]{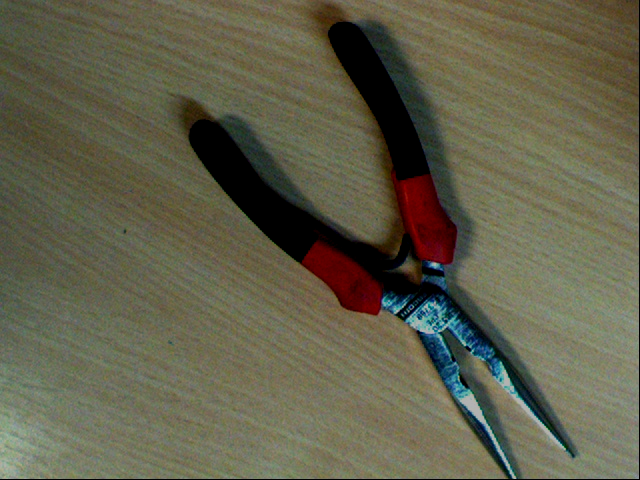}};
    \node[inner sep=0pt] (im0) at (.4\textwidth,0)
        {\includegraphics[width=.17\textwidth]{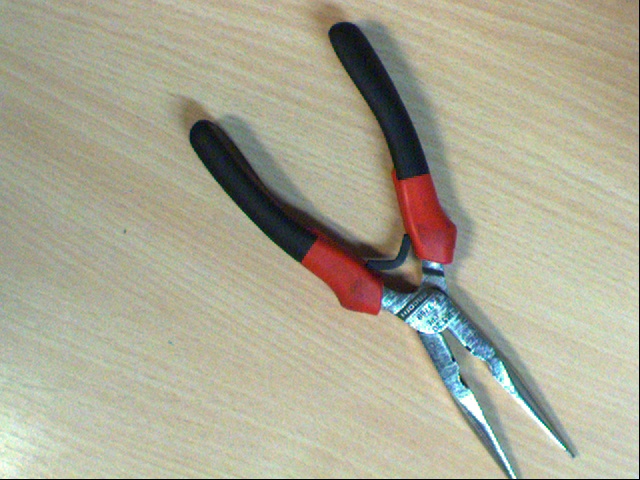}};
    \node[inner sep=0pt] (im0) at (.6\textwidth,0)
        {\includegraphics[width=.17\textwidth]{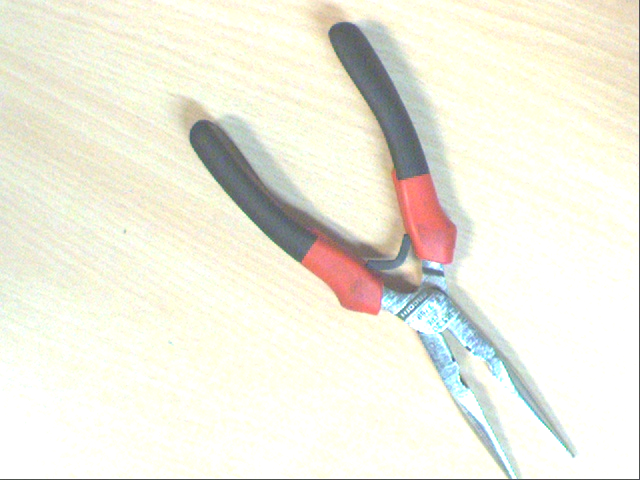}};
    \node[inner sep=0pt] (im0) at (.8\textwidth,0)
        {\includegraphics[width=.17\textwidth]{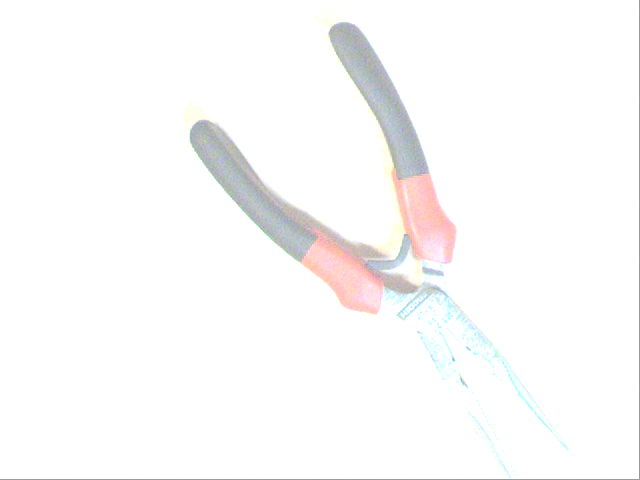}};
    \end{tikzpicture}
    
    \caption{Example images of the 5 artificial brightness conditions used to test different pipelines robustness.}
    \label{fig:brigthness}
\end{figure}

For further evaluation of the robustness to lighting conditions, we computationally modify the brightness of the pictures under conditions 2 only.  Example images with the applied brightness filters can be visualized on Figure \ref{fig:brigthness}. This allows to isolate the influence of brightness in the clustering results.

Then, for each BLC, we create a clustering problem by sampling one image (i.e. one position/orientation) for each object. To evaluate the 10 remaining pipelines (5 CNN architectures and 2 clustering algorithms), we sample 1000 clustering problems as explained above and try to cluster them with each of the pipelines. These experiments are used to choose the most robust clustering routine for our application.

\subsubsection{Results}

Results from the above experiments are reported in both Tables \ref{tab:robust_real} (physical BLC variation) and \ref{tab:robust_bright} (artificially modified brightness). Reported results are the averages over the 1000 runs. Both tables show that agglomerative clustering always present better results than KMeans. Also, Inception-like architectures tend to outperform their competitors, particularly Xception, which outperforms all the other CNN architectures for every clustering algorithm and every BLC. 

Based on these result, we choose Xception + AC as our full vision pipeline. This clustering routine is used in the next section in a framework making use of redundancy of images to improve URS robustness. The chosen pipeline seems to be robust to reasonable changes in lighting conditions (Dark and Bright) but its performances start to decrease when the brightness is really high or low. This makes sens as it also becomes difficult for a human to identify these objects. On the other hand, background changes seem to have a more important impact on the clustering results. Indeed, BLC4 shows much lower results than other BLC. This might come from the fact that the background in BLC4 contains geometric shapes (lines, circles) which are distractors for the network.

\input{table_single_robustness.tex}
\input{table_artificial_light.tex}

%% file: full_pipeline.tex
\newlength{\imwidth} \setlength{\imwidth}{0.15\textwidth}
\newlength{\xthr} \setlength{\xthr}{0.9cm}
\newlength{\ythr} \setlength{\ythr}{0.7cm}
\newlength{\yroww} \setlength{\yroww}{-2.5cm}
\newlength{\arrowlen} \setlength{\arrowlen}{0.1\textwidth}
\newlength{\xim} \setlength{\xim}{0.67\imwidth + \arrowlen}
\newlength{\yim} \setlength{\yim}{0cm}
\newlength{\dylab} \setlength{\dylab}{0.3cm}

\begin{tikzpicture}
    \node[inner sep=0pt] (scene) at (0,0) {\includegraphics[width=\imwidth]{inputs.jpg}};
    
    \node[inner sep=0pt] (im0) at (\xim, \yim + \ythr) {\includegraphics[width=0.35\imwidth]{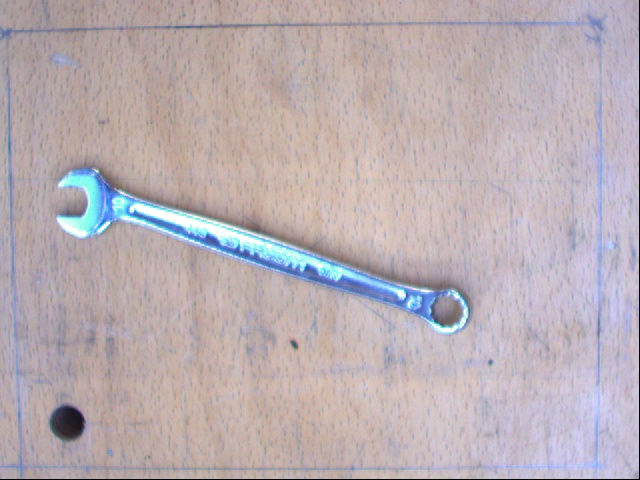}};
    \node[inner sep=0pt] (im1) at (\xim, \yim) {\includegraphics[width=0.35\imwidth]{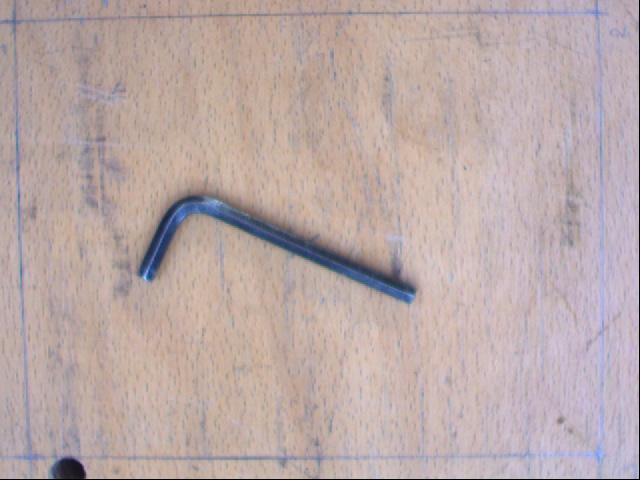}};
    \node[inner sep=0pt] (im2) at (\xim, \yim - \ythr) {\includegraphics[width=0.35\imwidth]{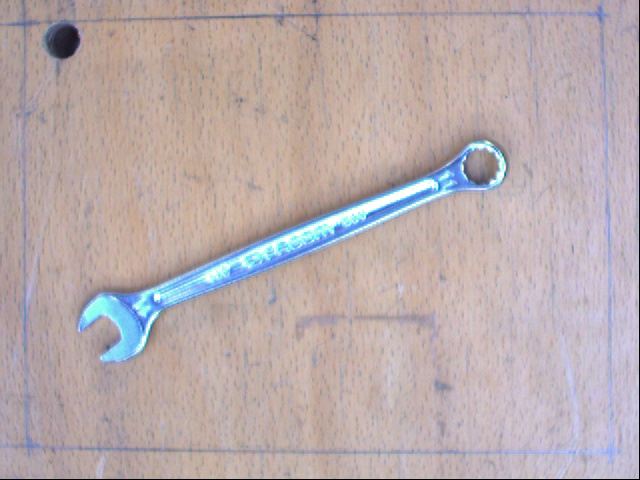}};
    \node[inner sep=0pt] (im3) at (\xim + \xthr, \yim + \ythr) {\includegraphics[width=0.35\imwidth]{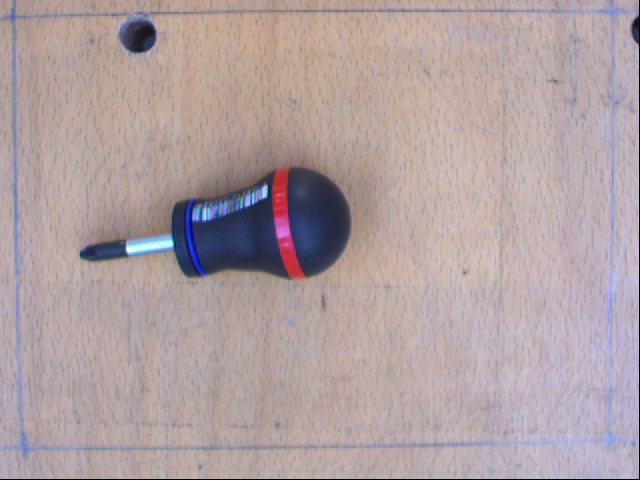}};
    \node[inner sep=0pt] (im4) at (\xim + \xthr, \yim) {\includegraphics[width=0.35\imwidth]{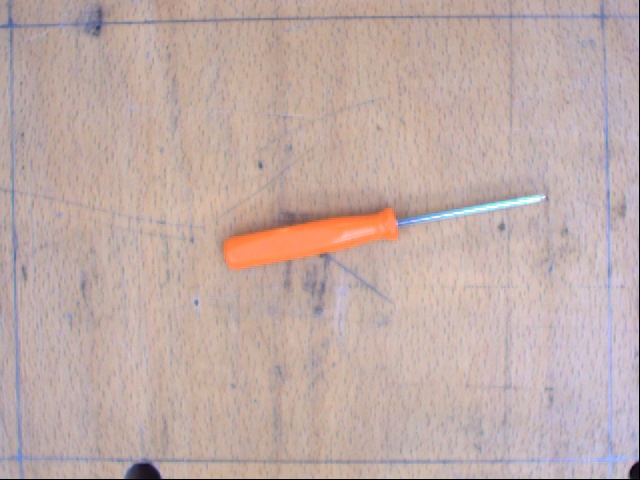}};
    \node[inner sep=0pt] (im5) at (\xim + \xthr, \yim - \ythr) {\includegraphics[width=0.35\imwidth]{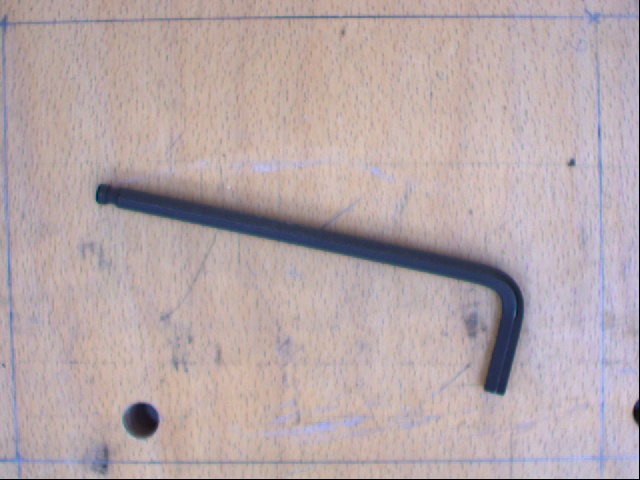}};
    \node[inner sep=0pt] (im6) at (\xim + 2 * \xthr, \yim + \ythr) {\includegraphics[width=0.35\imwidth]{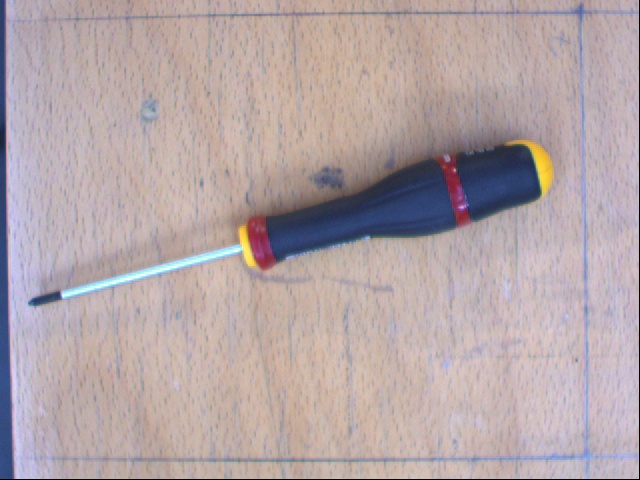}};
    \node[inner sep=0pt] (im7) at (\xim + 2 * \xthr, \yim) {\includegraphics[width=0.35\imwidth]{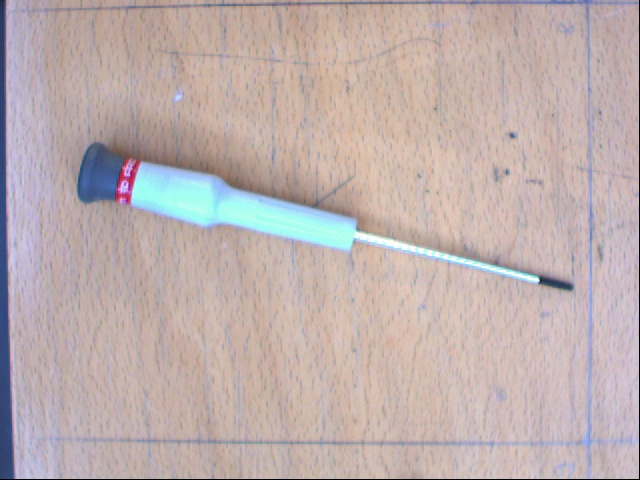}};
    \node[inner sep=0pt] (im8) at (\xim + 2 * \xthr, \yim - \ythr) {\includegraphics[width=0.35\imwidth]{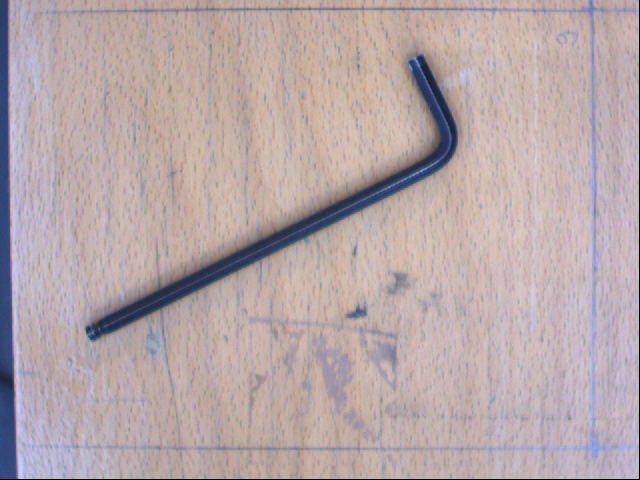}};
    
    {\larger[1]
    \node[inner sep=0pt, text width = \imwidth] (class) at (2.1\imwidth + 2\arrowlen, 0) {\textbf{$\{C(i)\}_{i \in Im}$}};}
    
    \node[inner sep=0pt] (out) at (3.35\imwidth + 3\arrowlen, 0) {\includegraphics[width=1.5\imwidth]{output.jpg}};
    
    \node[inner sep=0pt] (scan) at (0.5\imwidth + 0.5\arrowlen, \yroww) {\includegraphics[width=\imwidth]{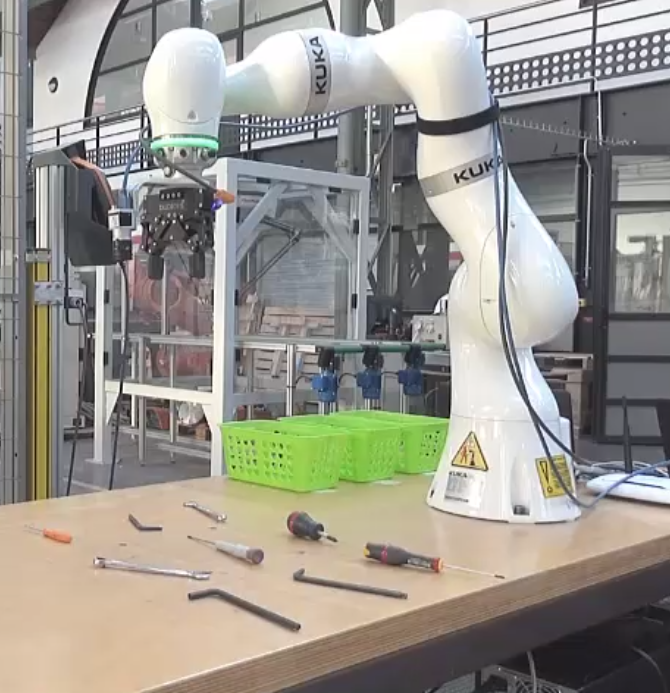}};
        
    {\larger[1] \node[rectangle, draw = black, rounded corners, minimum height=2em, fill = black!20] (cv) at (1.6\imwidth + 1.5\arrowlen, \yroww) {Vision Module};}
        
    \node[inner sep=0pt] (manip) at (2.6\imwidth + 2.5\arrowlen, \yroww) {\includegraphics[width=\imwidth]{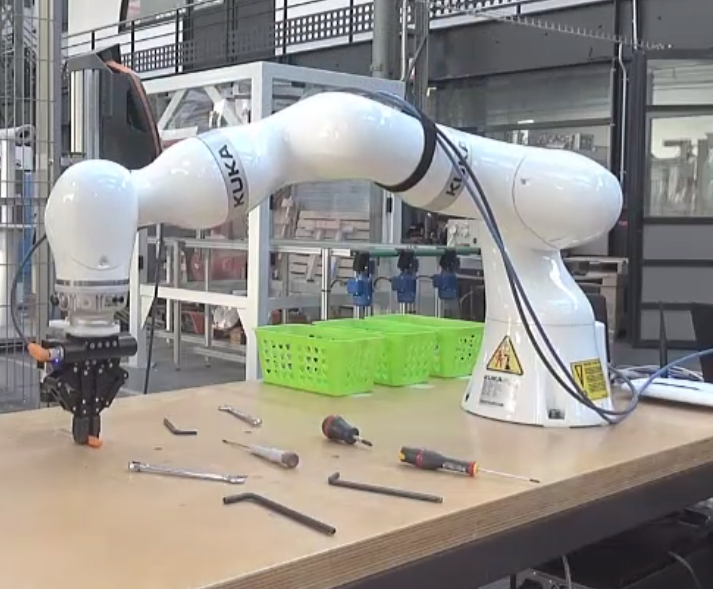}};
        
    \draw[->,line width=0.4mm] (scene.east) -- (im1.west);
    \draw[->,line width=0.4mm] (im7.east) -- (class.west);
    \draw[->,line width=0.4mm] (class.east) -- (out.west);
    
    \node[inner sep=0pt] (p1) at (0.5\imwidth + 0.5\arrowlen, 0) {};
    \node[inner sep=0pt] (p2) at (1.6\imwidth + 1.5\arrowlen, 0) {};
    \node[inner sep=0pt] (p3) at (2.6\imwidth + 2.5\arrowlen, 0) {};
    \draw (p1.center) -- (scan.north);
    \draw (p2.center) -- (cv.north);
    \draw (p3.center) -- (manip.north);
    
    \node[] (txtscene) at (0, 0.5\imwidth + \dylab) {\textit{Initial Scene}};
    \node[] (txtpic) at (\xim + \xthr, 0.5\imwidth + 0.5\dylab) {\textit{Image Set (Im)}};
    \node[] (txtclas) at (2.1\imwidth + 2\arrowlen, 0.5\imwidth - \dylab) {\textit{Category Set}};
    \node[] (txtout) at (3.35\imwidth + 3\arrowlen, 0.5\imwidth + 0.5\dylab) {\textit{Output}};
    \node[] (txtscan) at (0.5\imwidth + 0.5\arrowlen, \yroww - 0.5\imwidth - \dylab) {\textit{Scene Scanning}};
    \node[] (txtcv) at (1.6\imwidth + 1.5\arrowlen, \yroww - 0.5\imwidth + 0.5\dylab) {\textit{See Subfigure (\ref{subfig:vismod})}};
    \node[] (txtmanip) at (2.6\imwidth + 2.5\arrowlen, \yroww - 0.5\imwidth - 0.5\dylab) {\textit{Grasping and Sorting}};
    
\end{tikzpicture}

%% file: vision_unit.tex
\newcounter{numch1} \setcounter{numch1}{2}
\newcounter{numch2} \setcounter{numch2}{3}
\newcounter{numch3} \setcounter{numch3}{3}
    
\newlength{\xin} \setlength{\xin}{2cm}
\newlength{\yin} \setlength{\yin}{-0.5cm}
\newlength{\len} \setlength{\len}{1.2cm}
\newlength{\dec} \setlength{\dec}{-0.1cm}

\newlength{\xinn} \setlength{\xinn}{\xin + 1.8cm}
\newlength{\yinn} \setlength{\yinn}{\yin + 0.2cm}
\newlength{\lenn} \setlength{\lenn}{0.9cm}
\newlength{\decc} \setlength{\decc}{-0.1cm}

\newlength{\xinnn} \setlength{\xinnn}{\xin + 3.2cm}
\newlength{\yinnn} \setlength{\yinnn}{\yin + 0.3cm}
\newlength{\lennn} \setlength{\lennn}{0.8cm}
\newlength{\deccc} \setlength{\deccc}{-0.1cm}

\newlength{\lsq} \setlength{\lsq}{0.2cm}
\newlength{\lsqq} \setlength{\lsqq}{0.15cm}

\newcounter{numcir} \setcounter{numcir}{4}
\newlength{\cirsize} \setlength{\cirsize}{0.4cm}
\newlength{\xcirini} \setlength{\xcirini}{8.8cm}
\newlength{\ycir} \setlength{\ycir}{-0.7cm}
\newlength{\cirspace} \setlength{\cirspace}{0.5cm}
\newlength{\yspace} \setlength{\yspace}{0.5cm}
    
\begin{tikzpicture}
\centering
    \node[inner sep=0pt] (imset) at (0, 0) {};
    \node[inner sep=0pt] (xceini) at (\xin - 0.5cm, 0) {};
    
    \input{fig_xce.tex}
    
    \node[inner sep=0pt] (xcefin) at (\xinnn + \lennn + 0.8cm, 0) {};
    
    \node[inner sep=0pt] (aggini) at (\xcirini - 0.5cm, 0) {};
    
    \input{fig_agglo.tex}
    
    \node[inner sep=0pt] (aggfin) at (\xcirini + 4 * \cirspace + 0.5cm, 0) {};
    
    \node[inner sep=0pt] (class) at (\xcirini + 4 * \cirspace + 2cm, 0) {};
    
    \draw[->,line width=0.4mm] (imset.east) -- (xceini.west);
    \draw[->,line width=0.4mm] (xcefin.east) -- (aggini.west);
    \draw[->,line width=0.4mm] (aggfin.east) -- (class.west);
    
    \node[inner sep=0pt] (txtim) at (0.7, 0.4) {Images};
    \node[inner sep=0pt] (txtfv) at (7.5, 0.4) {Features};
    \node[inner sep=0pt] (txtcls) at (12, 0.4) {Classes};
    \node[inner sep=0pt] (txtxce) at (4.2, -1.5) {\textbf{CNN Feature Extraction}};
    \node[inner sep=0pt] (txtagg) at (10, -1.5) {\textbf{Standard Clustering}};
    \node[inner sep=0pt] (txtagg) at (10, -1.9) {\textbf{Algorithm}};
    
\end{tikzpicture}

%% file: fig_xce.tex
\foreach \i in {0, ..., \value{numch1}}
{
    \draw[fill=black,opacity=0.2,draw=black] (\xin + \i * \dec, \yin + \i * \dec) -- (\xin + \i * \dec + \len, \yin + \i * \dec) -- (\xin + \i * \dec + \len, \yin + \i * \dec + \len) -- (\xin + \i * \dec, \yin + \i * \dec + \len) -- (\xin + \i * \dec, \yin + \i * \dec);
}

\draw (\xin + 0.5 * \len, \yin + 0.5 * \len) -- (\xin + 0.5 * \len + \lsq, \yin + 0.5 * \len) -- (\xin + 0.5 * \len + \lsq, \yin + 0.5 * \len + \lsq) -- (\xin + 0.5 * \len, \yin + 0.5 * \len + \lsq) -- (\xin + 0.5 * \len, \yin + 0.5 * \len);

\draw (\xin + 0.5 * \len + \lsq, \yin + 0.5 * \len) -- (\xinn, \yin + 0.5 * \len + 0.5 * \lsq);
\draw (\xin + 0.5 * \len + \lsq, \yin + 0.5 * \len + \lsq) -- (\xinn, \yin + 0.5 * \len + 0.5 * \lsq);

\foreach \i in {0, ..., \value{numch2}}
{
    \draw[fill=black,opacity=0.2,draw=black] (\xinn + \i * \decc, \yinn + \i * \decc) -- (\xinn + \i * \decc + \lenn, \yinn + \i * \decc) -- (\xinn + \i * \decc + \lenn, \yinn + \i * \decc + \lenn) -- (\xinn + \i * \decc, \yinn + \i * \decc + \lenn) -- (\xinn + \i * \decc, \yinn + \i * \decc);
}

\draw (\xinn + 0.5 * \lenn, \yinn + 0.5 * \lenn) -- (\xinn + 0.5 * \lenn + \lsqq, \yinn + 0.5 * \lenn) -- (\xinn + 0.5 * \lenn + \lsqq, \yinn + 0.5 * \lenn + \lsqq) -- (\xinn + 0.5 * \lenn, \yinn + 0.5 * \lenn + \lsqq) -- (\xinn + 0.5 * \lenn, \yinn + 0.5 * \lenn);

\draw (\xinn + 0.5 * \lenn + \lsqq, \yinn + 0.5 * \lenn) -- (\xinnn, \yinn + 0.5 * \lenn + 0.5 * \lsqq);
\draw (\xinn + 0.5 * \lenn + \lsqq, \yinn + 0.5 * \lenn + \lsqq) -- (\xinnn, \yinn + 0.5 * \lenn + 0.5 * \lsqq);

\foreach \i in {0, ..., \value{numch3}}
{
    \draw[fill=black,opacity=0.2,draw=black] (\xinnn + \i * \deccc, \yinnn + \i * \deccc) -- (\xinnn + \i * \deccc + \lennn, \yinnn + \i * \deccc) -- (\xinnn + \i * \deccc + \lennn, \yinnn + \i * \deccc + \lennn) -- (\xinnn + \i * \deccc, \yinnn + \i * \deccc + \lennn) -- (\xinnn + \i * \deccc, \yinnn + \i * \deccc);
}

\draw (\xinnn + 1.5 * \lennn, \yinnn - 0.5 * \lennn) -- (\xinnn + 1.5 * \lennn + \lsq, \yinnn - 0.5 * \lennn) -- (\xinnn + 1.5 * \lennn + \lsq, \yinnn + 1.3 * \lennn) -- (\xinnn + 1.5 * \lennn, \yinnn + 1.3 * \lennn) -- (\xinnn + 1.5 * \lennn, \yinnn - 0.5 * \lennn);

\draw (\xinnn + \value{numch3} * \deccc + \lennn, \yinnn + \value{numch3} * \deccc + \lennn) -- (\xinnn + 1.5 * \lennn, \yinnn + 1.3 * \lennn);
\draw (\xinnn + \value{numch3} * \deccc + \lennn, \yinnn + \value{numch3} * \deccc) -- (\xinnn + 1.5 * \lennn, \yinnn - 0.5 * \lennn);
\draw (\xinnn + 0 * \deccc + \lennn, \yinnn + 0 * \deccc + \lennn) -- (\xinnn + 1.5 * \lennn, \yinnn + 1.3 * \lennn);
\draw (\xinnn + 0 * \deccc + \lennn, \yinnn + 0 * \deccc) -- (\xinnn + 1.5 * \lennn, \yinnn - 0.5 * \lennn);

\node[] at (\xinnn + 1.2 * \lennn, \yinnn - 0.8 * \lennn){\textit{\footnotesize Final layer}};

%% file: table_dataset.tex
\begin{table}[!ht]
\centering
\caption{Dataset statistics \textit{(BLC stands for Background/Lighting Conditions)}.}
\label{tab:new_datastat}

\begin{tabular}{c|c|c|c|c}
\#Images & Images size & \#Classes & \#Images per class & \#BLC \\ \hline
560 & 640 x 480 & 7 & 12 to 24 & 5
\end{tabular}

\end{table}

%% file: fig_imageSample.tex
\newlength{\wim} \setlength{\wim}{0.17\textwidth}
\newlength{\xdec} \setlength{\xdec}{0.18\textwidth}
\newlength{\ydec} \setlength{\ydec}{-0.14\textwidth}

\begin{figure}[!ht]
    \centering
    
    \begin{tikzpicture}
    
    \node[] (lab1) at (0, 0.5\wim) {BLC 1};
    \node[inner sep=0pt] (im0) at (0, 0)
        {\includegraphics[width=\wim] {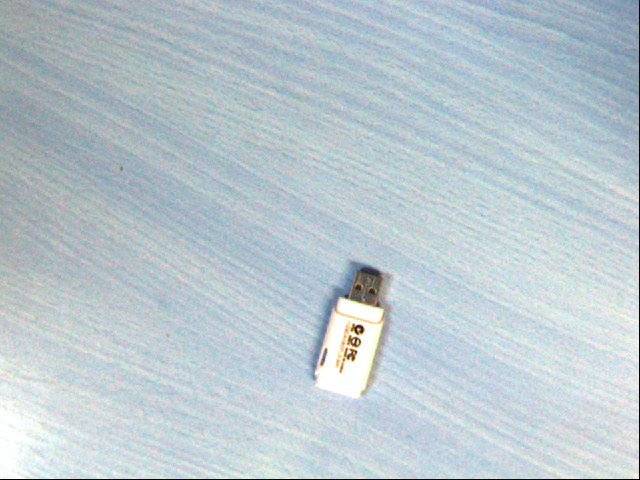}};
    \node[inner sep=0pt] (im1) at (0, \ydec)
        {\includegraphics[width=\wim] {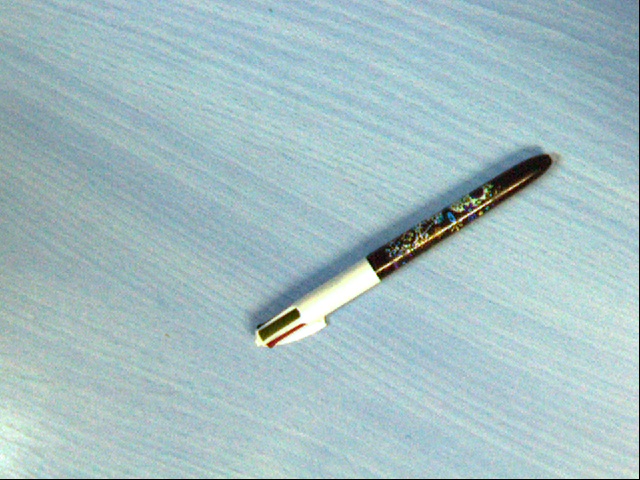}};
    \node[inner sep=0pt] (im2) at (0, 2\ydec)
        {\includegraphics[width=\wim] {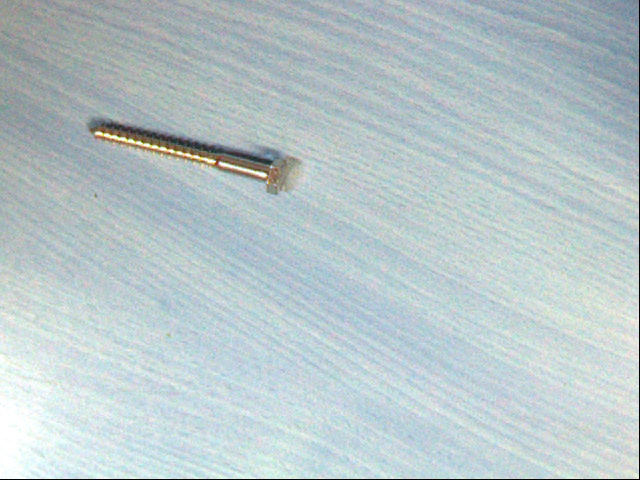}};
    
    \node[] (lab2) at (\xdec, 0.5\wim) {BLC 2};
    \node[inner sep=0pt] (im0) at (\xdec, 0)
        {\includegraphics[width=\wim] {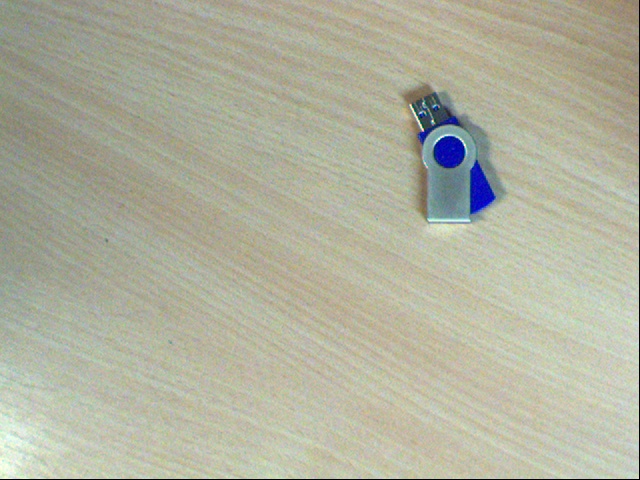}};
    \node[inner sep=0pt] (im1) at (\xdec, \ydec)
        {\includegraphics[width=\wim] {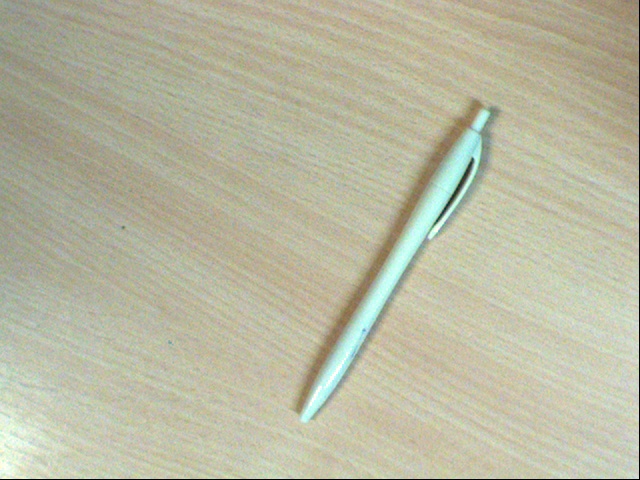}};
    \node[inner sep=0pt] (im2) at (\xdec, 2\ydec)
        {\includegraphics[width=\wim] {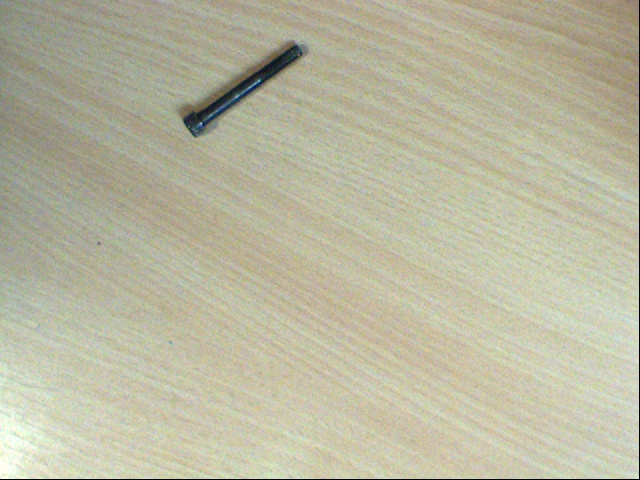}};
    
    \node[] (lab3) at (2\xdec, 0.5\wim) {BLC 3};
    \node[inner sep=0pt] (im0) at (2\xdec, 0)
        {\includegraphics[width=\wim] {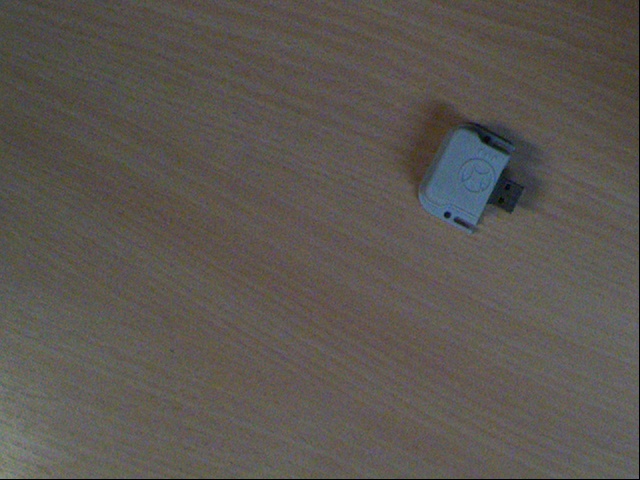}};
    \node[inner sep=0pt] (im1) at (2\xdec, \ydec)
        {\includegraphics[width=\wim] {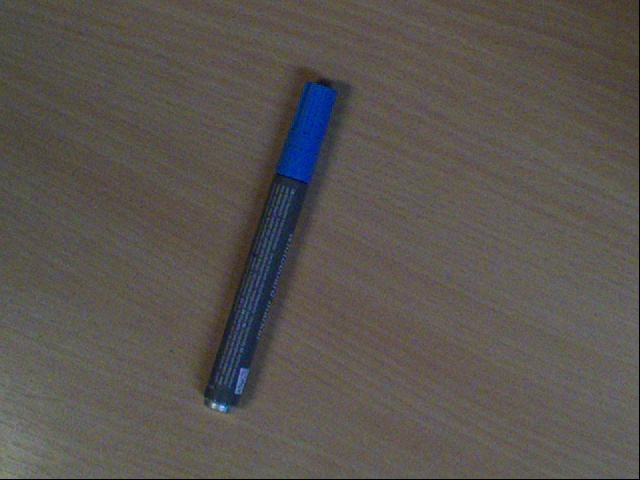}};
    \node[inner sep=0pt] (im2) at (2\xdec, 2\ydec)
        {\includegraphics[width=\wim] {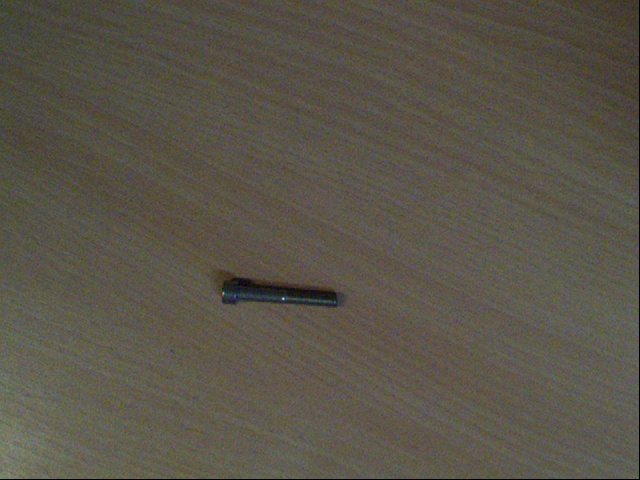}};

    \node[] (lab4) at (3\xdec, 0.5\wim) {BLC 4};
    \node[inner sep=0pt] (im0) at (3\xdec, 0)
        {\includegraphics[width=\wim] {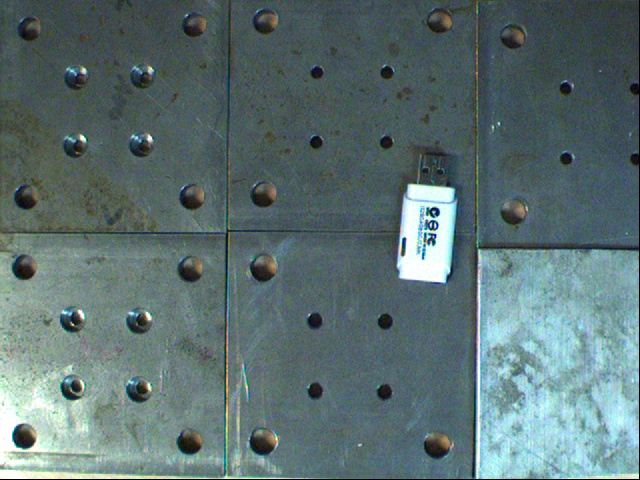}};
    \node[inner sep=0pt] (im1) at (3\xdec, \ydec)
        {\includegraphics[width=\wim] {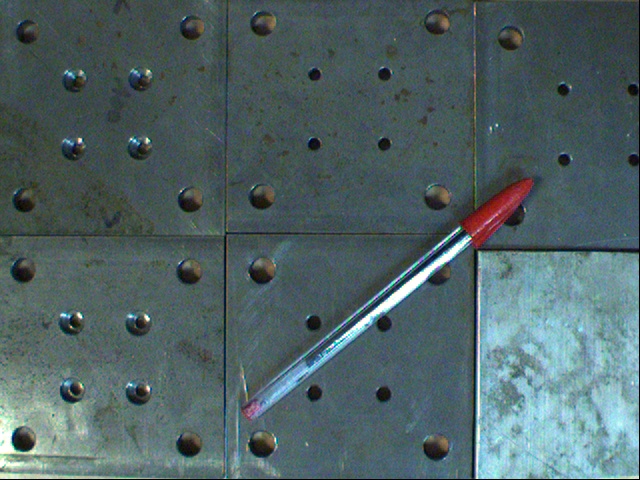}};
    \node[inner sep=0pt] (im2) at (3\xdec, 2\ydec)
        {\includegraphics[width=\wim] {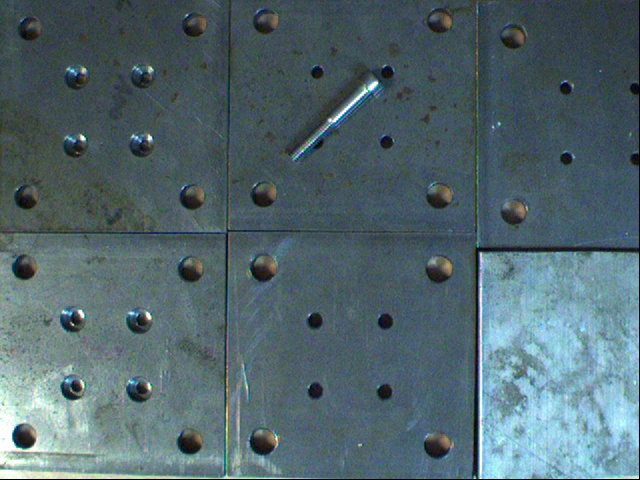}};

    \node[] (lab5) at (4\xdec, 0.5\wim) {BLC 5};
    \node[inner sep=0pt] (im0) at (4\xdec, 0)
        {\includegraphics[width=\wim] {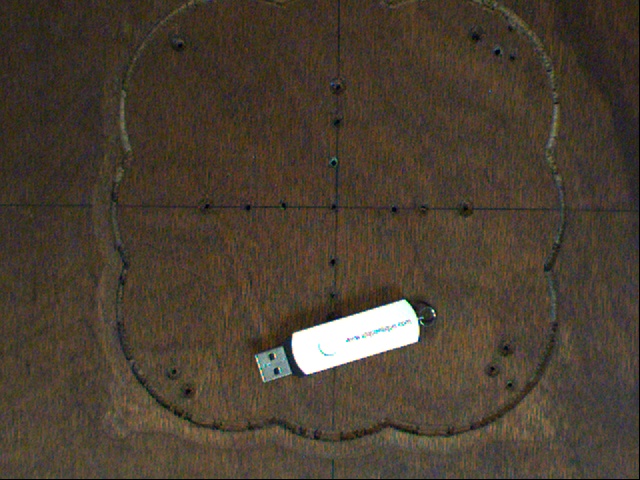}};
    \node[inner sep=0pt] (im1) at (4\xdec, \ydec)
        {\includegraphics[width=\wim] {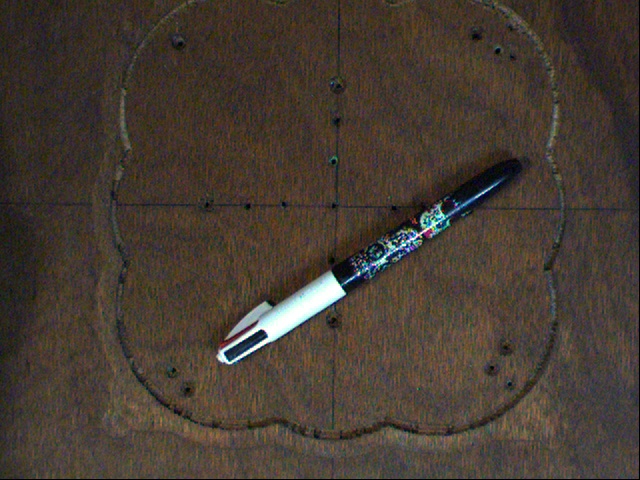}};
    \node[inner sep=0pt] (im2) at (4\xdec, 2\ydec)
        {\includegraphics[width=\wim] {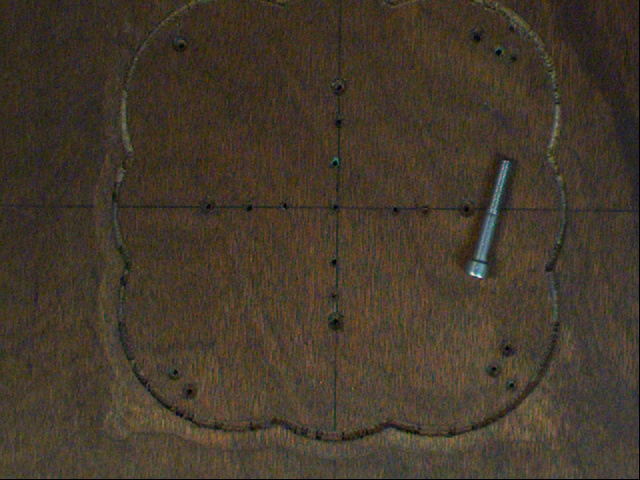}};

    \node[] (row1) at (-0.75\wim, 0) {USB}; 
    \node[] (row2) at (-0.75\wim, \ydec) {Pen};
    \node[] (row3) at (-0.75\wim, 2\ydec) {Screw};
    
    \end{tikzpicture}
    
    \caption{Sample images from the robustness validation dataset.}
    \label{fig:dataset}
\end{figure}

%% file: table_single_robustness.tex
\begin{table}[!ht]
\centering
\caption{Clustering results for different CNN architectures and clustering algorithms on the tool clustering dataset. NMI scores are in black and cluster purity in blue, italics.}
\label{tab:robust_real}

\vspace{\baselineskip}
    \scalebox{0.9}{
    \begin{tabular}{cc"c|c|c|c|c}
        
        \multicolumn{2}{c"}{} & BLC1 & BLC2 & BLC3 & BLC4 & BLC5 \tabularnewline
        \Xhline{2\arrayrulewidth}
        
        \multirow{4}{*}{Inception V3} & \multirow{2}{*}{AC} & 0.82 & 0.82 & 0.80 & 0.65 & 0.79 \tabularnewline 
        & & \textit{\textcolor{Blue}{0.81}} & \textit{\textcolor{Blue}{0.81}} & \textit{\textcolor{Blue}{0.80}} & \textit{\textcolor{Blue}{0.65}} & \textit{\textcolor{Blue}{0.76}} \tabularnewline \cline{2-7}
        & \multirow{2}{*}{KMeans} & 0.80 & 0.79 & 0.76 & 0.63 & 0.75 \tabularnewline 
        & & \textit{\textcolor{Blue}{0.79}} & \textit{\textcolor{Blue}{0.78}} & \textit{\textcolor{Blue}{0.76}} & \textit{\textcolor{Blue}{0.64}} & \textit{\textcolor{Blue}{0.73}} \tabularnewline \hline
        
        \multirow{4}{*}{Resnet 50} & \multirow{2}{*}{AC} & 0.81 & 0.74 & 0.74 & 0.62 & 0.72 \tabularnewline 
        & & \textit{\textcolor{Blue}{0.81}} & \textit{\textcolor{Blue}{0.74}} & \textit{\textcolor{Blue}{0.75}} & \textit{\textcolor{Blue}{0.59}} & \textit{\textcolor{Blue}{0.71}} \tabularnewline \cline{2-7}
        & \multirow{2}{*}{KMeans} & 0.77 & 0.71 & 0.71 & 0.58 & 0.70 \tabularnewline 
        & & \textit{\textcolor{Blue}{0.78}} & \textit{\textcolor{Blue}{0.72}} & \textit{\textcolor{Blue}{0.72}} & \textit{\textcolor{Blue}{0.58}} & \textit{\textcolor{Blue}{0.70}} \tabularnewline \hline
        
        \multirow{4}{*}{VGG 16} & \multirow{2}{*}{AC} & 0.76 & 0.74 & 0.73 & 0.61 & 0.70 \tabularnewline 
        & & \textit{\textcolor{Blue}{0.75}} & \textit{\textcolor{Blue}{0.73}} & \textit{\textcolor{Blue}{0.72}} & \textit{\textcolor{Blue}{0.60}} & \textit{\textcolor{Blue}{0.69}} \tabularnewline \cline{2-7}
        & \multirow{2}{*}{KMeans} & 0.72 & 0.70 & 0.71 & 0.58 & 0.67 \tabularnewline 
        & & \textit{\textcolor{Blue}{0.72}} & \textit{\textcolor{Blue}{0.70}} & \textit{\textcolor{Blue}{0.70}} & \textit{\textcolor{Blue}{0.57}} & \textit{\textcolor{Blue}{0.67}} \tabularnewline \hline
        
        \multirow{4}{*}{VGG 19} & \multirow{2}{*}{AC} & 0.76 & 0.77 & 0.71 & 0.59 & 0.71 \tabularnewline 
        & & \textit{\textcolor{Blue}{0.76}} & \textit{\textcolor{Blue}{0.76}} & \textit{\textcolor{Blue}{0.72}} & \textit{\textcolor{Blue}{0.58}} & \textit{\textcolor{Blue}{0.70}} \tabularnewline \cline{2-7}
        & \multirow{2}{*}{KMeans} & 0.73 & 0.73 & 0.69 & 0.56 & 0.67 \tabularnewline 
        & & \textit{\textcolor{Blue}{0.73}} & \textit{\textcolor{Blue}{0.73}} & \textit{\textcolor{Blue}{0.70}} & \textit{\textcolor{Blue}{0.56}} & \textit{\textcolor{Blue}{0.67}} \tabularnewline \hline
        
        \multirow{4}{*}{Xception} & \multirow{2}{*}{AC} & 0.86 & 0.90 & 0.84 & 0.69 & 0.83 \tabularnewline 
        & & \textit{\textcolor{Blue}{0.85}} & \textit{\textcolor{Blue}{0.90}} & \textit{\textcolor{Blue}{0.85}} & \textit{\textcolor{Blue}{0.69}} & \textit{\textcolor{Blue}{0.81}} \tabularnewline \cline{2-7}
        & \multirow{2}{*}{KMeans} & 0.84 & 0.87 & 0.82 & 0.66 & 0.80 \tabularnewline 
        & & \textit{\textcolor{Blue}{0.83}} & \textit{\textcolor{Blue}{0.86}} & \textit{\textcolor{Blue}{0.82}} & \textit{\textcolor{Blue}{0.66}} & \textit{\textcolor{Blue}{0.80}} \tabularnewline

    \end{tabular}}
    
\end{table}

%% file: table_artificial_light.tex
\begin{table}[!ht]
\centering
\caption{Clustering results for different CNN architectures and clustering algorithms for different artificially modified lighting conditions on the BLC2 subset. NMI scores are in black and cluster purity in blue, italics.}
\label{tab:robust_bright}

\vspace{\baselineskip}
    \scalebox{0.9}{
    \begin{tabular}{cc"c|c|c|c|c}
        
        \multicolumn{2}{c"}{} & Very dark & Dark & Normal & Bright & Very bright \tabularnewline
        \Xhline{2\arrayrulewidth}
        \multirow{4}{*}{Inception V3} & \multirow{2}{*}{AC} & 0.74 & 0.81 & 0.82 & 0.80 & 0.71 \tabularnewline 
        & & \textit{\textcolor{Blue}{0.72}} & \textit{\textcolor{Blue}{0.79}} & \textit{\textcolor{Blue}{0.81}} & \textit{\textcolor{Blue}{0.80}} & \textit{\textcolor{Blue}{0.70}}\tabularnewline \cline{2-7}
        & \multirow{2}{*}{KMeans} & 0.70 & 0.77 & 0.79 & 0.77 & 0.66 \tabularnewline 
        & & \textit{\textcolor{Blue}{0.70}} & \textit{\textcolor{Blue}{0.75}} & \textit{\textcolor{Blue}{0.78}} & \textit{\textcolor{Blue}{0.77}} & \textit{\textcolor{Blue}{0.67}} \tabularnewline \hline

        \multirow{4}{*}{Resnet 50} & \multirow{2}{*}{AC} & 0.67 & 0.73 & 0.74 & 0.69 & 0.61 \tabularnewline 
        & & \textit{\textcolor{Blue}{0.67}} & \textit{\textcolor{Blue}{0.73}} & \textit{\textcolor{Blue}{0.74}} & \textit{\textcolor{Blue}{0.68}} & \textit{\textcolor{Blue}{0.61}}\tabularnewline \cline{2-7}
        & \multirow{2}{*}{KMeans} & 0.65 & 0.70 & 0.71 & 0.66 & 0.58 \tabularnewline 
        & & \textit{\textcolor{Blue}{0.66}} & \textit{\textcolor{Blue}{0.71}} & \textit{\textcolor{Blue}{0.72}} & \textit{\textcolor{Blue}{0.66}} & \textit{\textcolor{Blue}{0.59}} \tabularnewline \hline

        \multirow{4}{*}{VGG 16} & \multirow{2}{*}{AC} & 0.66 & 0.73 & 0.74 & 0.68 & 0.61 \tabularnewline 
        & & \textit{\textcolor{Blue}{0.66}} & \textit{\textcolor{Blue}{0.72}} & \textit{\textcolor{Blue}{0.73}} & \textit{\textcolor{Blue}{0.68}} & \textit{\textcolor{Blue}{0.61}}\tabularnewline \cline{2-7}
        & \multirow{2}{*}{KMeans} & 0.62 & 0.69 & 0.70 & 0.65 & 0.57 \tabularnewline 
        & & \textit{\textcolor{Blue}{0.63}} & \textit{\textcolor{Blue}{0.69}} & \textit{\textcolor{Blue}{0.70}} & \textit{\textcolor{Blue}{0.66}} & \textit{\textcolor{Blue}{0.58}} \tabularnewline \hline

        \multirow{4}{*}{VGG 19} & \multirow{2}{*}{AC} & 0.67 & 0.76 & 0.77 & 0.74 & 0.64 \tabularnewline 
        & & \textit{\textcolor{Blue}{0.67}} & \textit{\textcolor{Blue}{0.75}} & \textit{\textcolor{Blue}{0.76}} & \textit{\textcolor{Blue}{0.72}} & \textit{\textcolor{Blue}{0.65}}\tabularnewline \cline{2-7}
        & \multirow{2}{*}{KMeans} & 0.64 & 0.73 & 0.73 & 0.71 & 0.59 \tabularnewline 
        & & \textit{\textcolor{Blue}{0.65}} & \textit{\textcolor{Blue}{0.73}} & \textit{\textcolor{Blue}{0.73}} & \textit{\textcolor{Blue}{0.70}} & \textit{\textcolor{Blue}{0.62}} \tabularnewline \hline

        \multirow{4}{*}{Xception} & \multirow{2}{*}{AC} & 0.77 & 0.88 & 0.90 & 0.84 & 0.73 \tabularnewline 
        & & \textit{\textcolor{Blue}{0.77}} & \textit{\textcolor{Blue}{0.89}} & \textit{\textcolor{Blue}{0.90}} & \textit{\textcolor{Blue}{0.84}} & \textit{\textcolor{Blue}{0.74}}\tabularnewline \cline{2-7}
        & \multirow{2}{*}{KMeans} & 0.74 & 0.85 & 0.87 & 0.82 & 0.70 \tabularnewline 
        & & \textit{\textcolor{Blue}{0.74}} & \textit{\textcolor{Blue}{0.86}} & \textit{\textcolor{Blue}{0.86}} & \textit{\textcolor{Blue}{0.82}} & \textit{\textcolor{Blue}{0.71}} \tabularnewline 
        
    \end{tabular}}
    
\end{table}

%% file: multiview.tex
\section{Increase robustness using multiple views}
\label{sec:multiview}

In the previous sections, for each object, we only use one image. However, URS deals with physical objects and does not need to be limited to single-image clustering. Using multiple views can increase the performance of the application by using redundant and complementary information. Hence, we propose a multi-view clustering method to solve the URS problem more accurately.

\subsection{Multi-view clustering using Ensemble-clustering}

Let $\Omega = \{\Omega_{1}, ... \Omega_{n}\}$ be the set of $n$ objects to cluster. For each object $\Omega_i$, let $\omega_i = \{\omega_{i, 1}, ... \omega_{i, m_i}\}$ be the set of $m_i$ views representing it. And let $C$ denote the clustering pipeline chosen from the previous experiments (Xception + AC). The proposed Multi-View Ensemble Clustering (MVEC) method, inspired by \cite{multiview_ensemble}, goes as follows. For each object $\Omega_i$, we randomly select one image $\omega_{i, j}$. Then, we cluster the new image set using $C$, the cluster assignments (partition) generated, is denoted $P_k$. This procedure is repeated $N$ times and we note $P = \{P_1, ... P_N\}$ the set of $N$ partitions.

Once the $N$ partitions have been generated  by randomly sampling images in each $\omega_i$, MVEC must find a consensus partition $P^*$, maximizing the agreement between the different partitions. In this paper, we choose an approach based on objects co-occurrence to compute $P^*$. Since we deal with unsupervised classification, there is no correspondence between the class assignments of the different partitions of $P$. To avoid this problem, we use an intermediate representation of $P$, called the co-association matrix: $CA$. $CA$ is a $n \times n$ matrix which entries are defined by
\begin{equation}
\label{eq:ca}
    CA_{pq} = \frac{1}{N} \sum_{t=1}^{N} \delta(P_t(\Omega_p), P_t(\Omega_q)),
\end{equation}
where $P_t(\Omega_i)$ is the label associated with object $\Omega_i$ and generated by partition $P_t$, and $\delta(a, b)$ is the Kronecker symbol, which is $1$ if $a=b$ and $0$ otherwise. Entry $(i,j)$ of $CA$ measures how many times objects $\Omega_i$ and $\Omega_j$ have been classified together by the different partitions.

Finally, let $C^*$ be any connectivity-based clustering algorithm. For instance, $C^*$ can be variants of agglomerative clustering or spectral clustering. The consensus partition is obtain by applying $C^*$ using $CA$ as the precomputed similarity measure between objects.
\begin{equation}
\label{eq:cons}
    P^* = C^*(CA).    
\end{equation}

A schematic representation for the MVEC pipeline can be found in Figure \ref{fig:multiview}. In our implementation, image sampling for partition generation is uniform, $C^*$ is agglomerative clustering and the number of partitions generated is $N = 1000$.

\input{multiview_method.tex}

\subsection{Results}

The results using MVEC with the clustering pipeline chosen above are reported in Table \ref{tab:multiview}. MVEC consistently performs significantly better than the corresponding single view approach (written in parenthesis in Table \ref{tab:multiview} as a reminder). It is more robust to poor background (BLC4) and lighting (Very bright) conditions. 

The excellent results obtained by MVEC were expectable. Indeed, by increasing the number of views, we introduce complementary information available for clustering. MVEC is naturally more robust to the unlucky event of selecting a poor view. Robustness to poor lighting conditions also makes sense as light comes from a certain direction and there are always better angles to observe the objects. Moreover, the same object, observed from different angles, can look very different, hence, by increasing the number of views, we also make more likely the event of having similar images for similar objects.

Regarding computation time, although MVEC takes longer, this time can be drastically reduced by parallelizing both the partition generation process and the co-association matrix computation.

\input{table_multi_robustness.tex}

%% file: multiview_method.tex
\newlength{\layerspace} \setlength{\layerspace}{0.23\textwidth}
\newlength{\inpxspace} \setlength{\inpxspace}{25 pt}
\newlength{\inpyspace} \setlength{\inpyspace}{22 pt}
\newlength{\layspace} \setlength{\layspace}{150 pt}

\begin{figure}
    \centering
    
        \begin{tikzpicture}
            \node[inner sep=0pt] (im11) at (0,0) {\includegraphics[width=.05\textwidth]{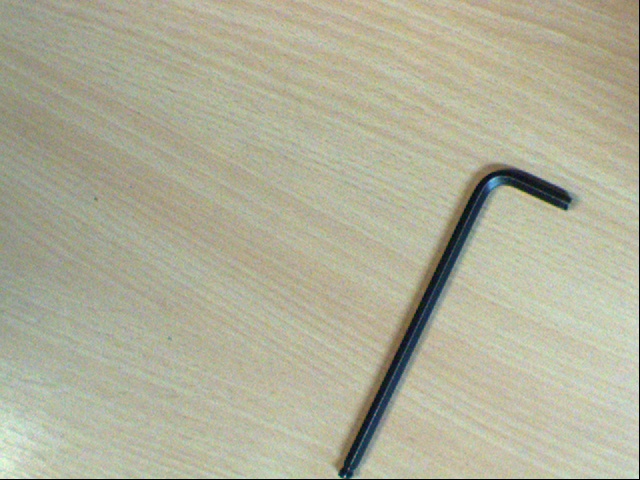}};
            \node[inner sep=0pt] (im12) at (\inpxspace,0) {\includegraphics[width=.05\textwidth]{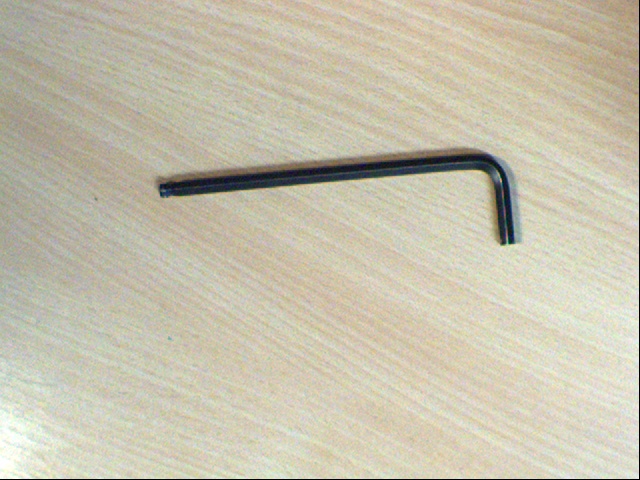}};
            \node[inner sep=0pt] (im13) at (2\inpxspace,0) {\includegraphics[width=.05\textwidth]{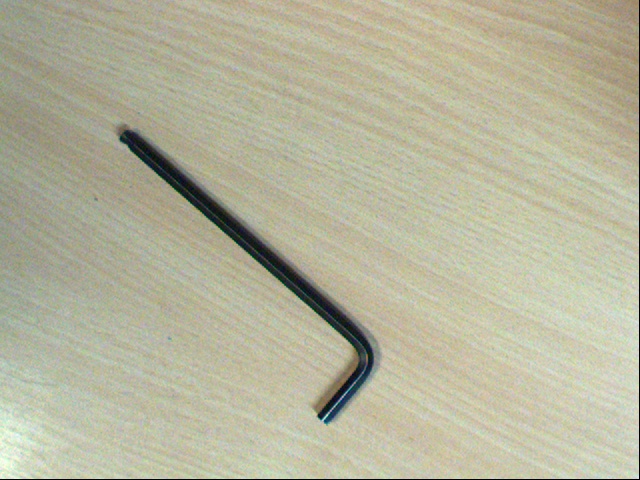}};
            \node[inner sep=0pt] (im14) at (3\inpxspace,0) {\includegraphics[width=.05\textwidth]{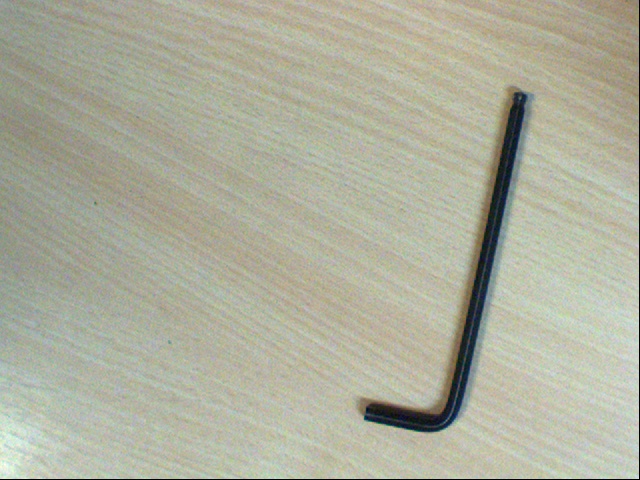}};
            \node[inner sep=0pt] (im21) at (0,2\inpyspace) {\includegraphics[width=.05\textwidth]{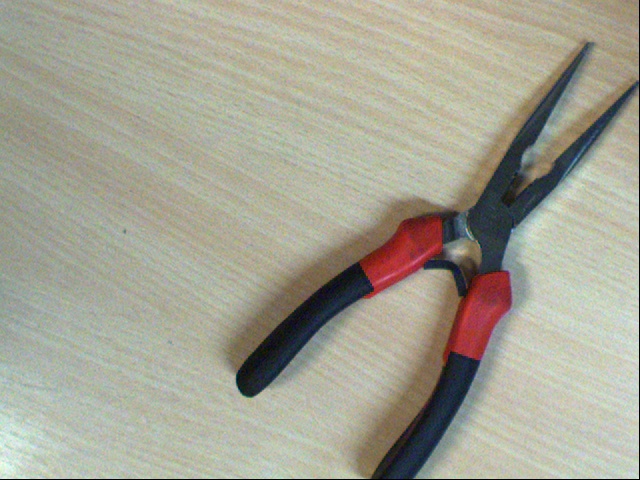}};
            \node[inner sep=0pt] (im22) at (1\inpxspace,2\inpyspace) {\includegraphics[width=.05\textwidth]{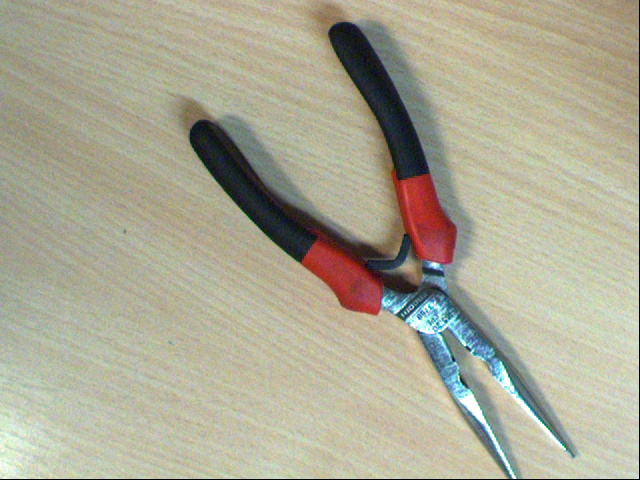}};
            \node[inner sep=0pt] (im23) at (2\inpxspace,2\inpyspace) {\includegraphics[width=.05\textwidth]{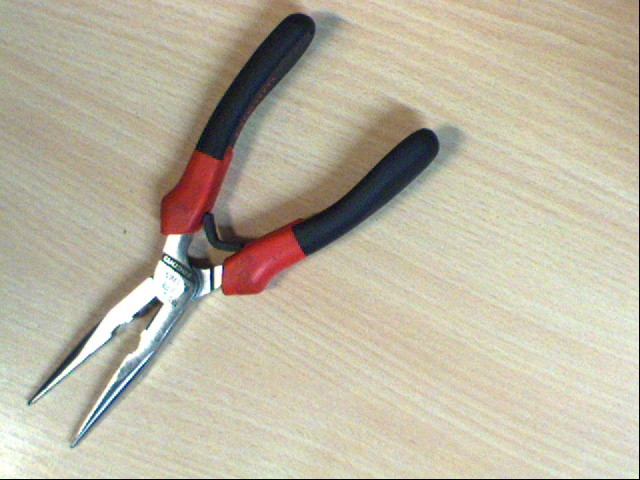}};
            \node[inner sep=0pt] (im24) at (3\inpxspace,2\inpyspace) {\includegraphics[width=.05\textwidth]{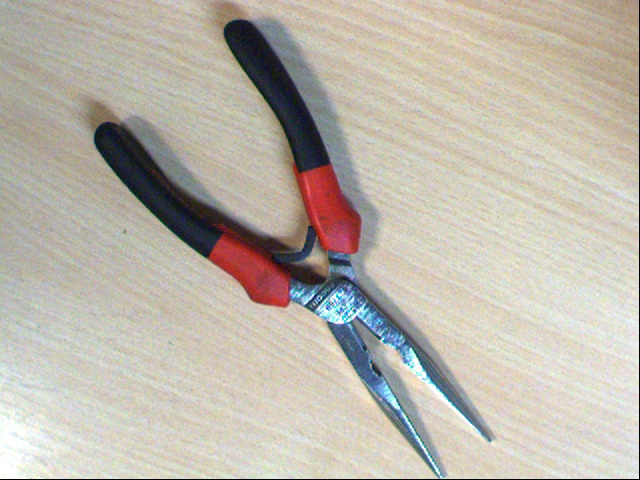}};
            \node[inner sep=0pt] (im31) at (0,3\inpyspace) {\includegraphics[width=.05\textwidth]{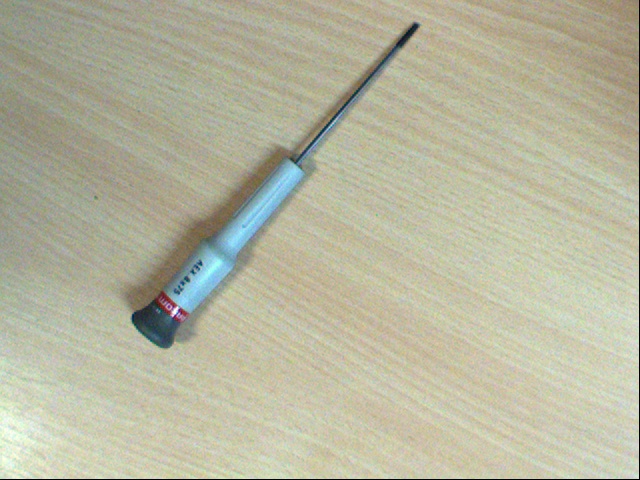}};
            \node[inner sep=0pt] (im32) at (1\inpxspace,3\inpyspace) {\includegraphics[width=.05\textwidth]{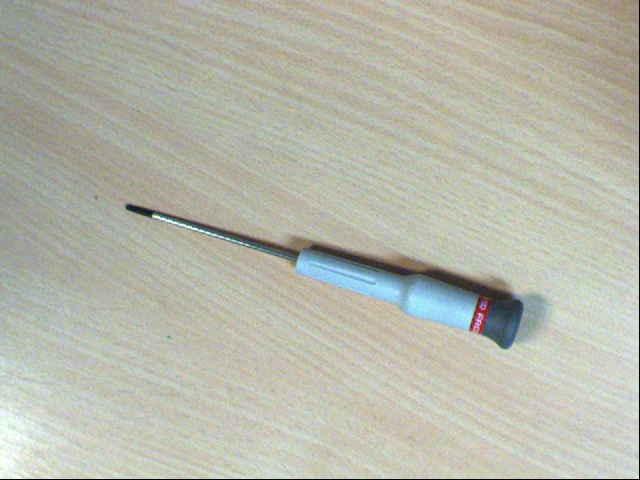}};
            \node[inner sep=0pt] (im33) at (2\inpxspace,3\inpyspace) {\includegraphics[width=.05\textwidth]{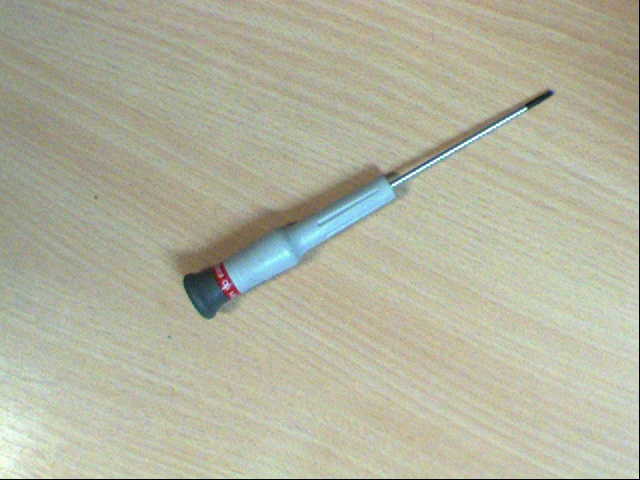}};
            \node[inner sep=0pt] (im34) at (3\inpxspace,3\inpyspace) {\includegraphics[width=.05\textwidth]{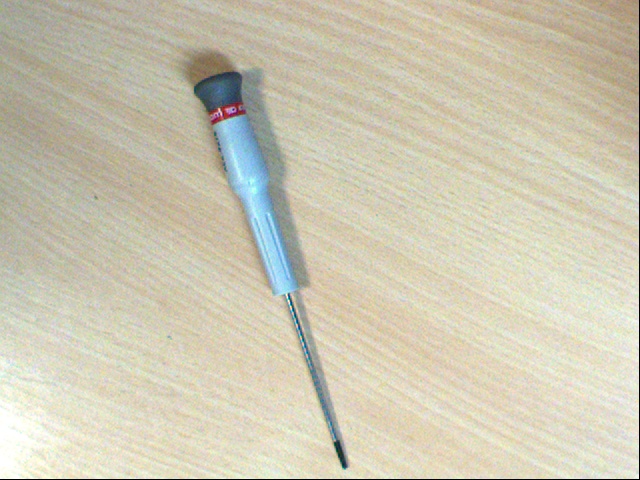}};
            \node[inner sep=0pt] (im_txt) at (1.5\inpxspace, 4.5\inpyspace) {\textit{Multiview Input}};
            \node[inner sep=0pt] (im_txt) at (1.5\inpxspace, 4\inpyspace) {\textit{images}};
            \node[inner sep=0pt] (dots1) at (1.5\inpxspace, \inpyspace) {$\dots$};
            \draw ($(im34.north east)+(0.1,0.1)$) rectangle ($(im11.south west)+(-0.1,-0.1)$);
            \draw ($(im34.north east)+(0.1,0.1)$) rectangle ($(im31.south west)+(-0.1,-0.1)$);
            \draw ($(im24.north east)+(0.1,0.1)$) rectangle ($(im21.south west)+(-0.1,-0.1)$);
            \draw ($(im24.south east)+(0.1,-0.1)$) rectangle ($(im11.north west)+(-0.1,0.1)$);
            \draw ($(im14.north east)+(0.1,0.1)$) rectangle ($(im11.south west)+(-0.1,-0.1)$);
            
            \node[inner sep=0pt] (rvs_txt1) at (0.72\layspace, -1.5\inpyspace) {\textit{Random}};
            \node[inner sep=0pt] (rvs_txt2) at (0.72\layspace, -2\inpyspace) {\textit{view sampling}};
            
            \node[inner sep=0pt] (clu1) at (\layspace, 3.5\inpyspace)
            {FE + C};
            \draw ($(clu1.north east)+(0.1,0.1)$) rectangle ($(clu1.south west)+(-0.1,-0.1)$);
            \node[inner sep=0pt] (clu2) at (\layspace, 2.5\inpyspace)
            {FE + C};
            \draw ($(clu2.north east)+(0.1,0.1)$) rectangle ($(clu2.south west)+(-0.1,-0.1)$);
            \node[inner sep=0pt] (clu3) at (\layspace, -0.5\inpyspace)
            {FE + C};
            \draw ($(clu3.north east)+(0.1,0.1)$) rectangle ($(clu3.south west)+(-0.1,-0.1)$);
            \path (clu2) -- (clu3) node [font=\Large, midway, sloped] {$\dots$};
            \node[inner sep=0pt] (clu_txt) at (\layspace, 4.9\inpyspace) {\textit{Image clustering}};
            \node[inner sep=0pt] (clu_txt) at (\layspace, 4.4\inpyspace) {\textit{pipeline}};
            
            \node[inner sep=0pt] (cam) at (2.6\layerspace, 1.5\inpyspace)
            {\includegraphics[width=.17\textwidth, height=.17\textwidth]{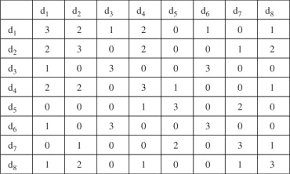}};
            \draw (cam.north east) rectangle (cam.south west);
            \node[inner sep=0pt] (coas_txt) at (2.6\layerspace, 4.2\inpyspace) {\textit{Co-Association}};
            \node[inner sep=0pt] (coas_txt) at (2.6\layerspace, 3.7\inpyspace) {\textit{Matrix}};
            
            \node[inner sep=0pt] (lab_txt1) at (2.3\layspace, 1.8\inpyspace) {\textbf{Final Set}};
            \node[inner sep=0pt] (lab_txt2) at (2.3\layspace, 1.2\inpyspace) {\textbf{of Labels}};
            
            \node[inner sep=0pt] (pg_txt1) at (1.3\layspace, -1.2\inpyspace) {\textit{Partition}};
            \node[inner sep=0pt] (pg_txt2) at (1.3\layspace, -1.7\inpyspace) {\textit{gathering}};
            
            \node[inner sep=0pt] (fc_txt) at (2.04\layspace, -1\inpyspace) {\textit{Final clustering}};
            
            \draw[->, line width = 0.3mm] ($(im34.east)+(0.1,0)$) -- ($(clu1.west)+(-0.15,0)$);
            \draw[->, line width = 0.3mm] ($(im24.east)+(0.1,0)$) -- ($(clu1.west)+(-0.15,0)$);
            \draw[->, line width = 0.3mm] ($(im14.east)+(0.1,0)$) -- ($(clu1.west)+(-0.15,0)$);
            \draw[->, line width = 0.3mm] ($(im34.east)+(0.1,0)$) -- ($(clu2.west)+(-0.15,0)$);
            \draw[->, line width = 0.3mm] ($(im24.east)+(0.1,0)$) -- ($(clu2.west)+(-0.15,0)$);
            \draw[->, line width = 0.3mm] ($(im14.east)+(0.1,0)$) -- ($(clu2.west)+(-0.15,0)$);
            \draw[->, line width = 0.3mm] ($(im34.east)+(0.1,0)$) -- ($(clu3.west)+(-0.15,0)$);
            \draw[->, line width = 0.3mm] ($(im24.east)+(0.1,0)$) -- ($(clu3.west)+(-0.15,0)$);
            \draw[->, line width = 0.3mm] ($(im14.east)+(0.1,0)$) -- ($(clu3.west)+(-0.15,0)$);
            \draw[->, line width = 0.3mm] ($(clu1.east)+(0.1,0)$) -- ($(cam.west)+(-0.15,0)$);
            \draw[->, line width = 0.3mm] ($(clu2.east)+(0.1,0)$) -- ($(cam.west)+(-0.15,0)$);
            \draw[->, line width = 0.3mm] ($(clu3.east)+(0.1,0)$) -- ($(cam.west)+(-0.15,0)$);
            \draw[->, line width = 0.3mm] ($(cam.east)+(0,0)$) -- ($(lab_txt2.north west)+(0,0.08)$);
            
            \draw[line width = 0.15mm] ($(im14.south east)+(0.78,0)$) -- ($(rvs_txt1.north)+(0,0.1)$);
            \draw[line width = 0.15mm] ($(clu3.north east)+(0.9,0.4)$) -- ($(pg_txt1.north)+(0,0.1)$);
            \draw[line width = 0.15mm] ($(cam.east)+(0.5,-0.1)$) -- ($(fc_txt.north)+(0,0.1)$);
        \end{tikzpicture}
    
    \caption{Proposed MVEC approach to use multiple views of each object.}
    \label{fig:multiview}
\end{figure}
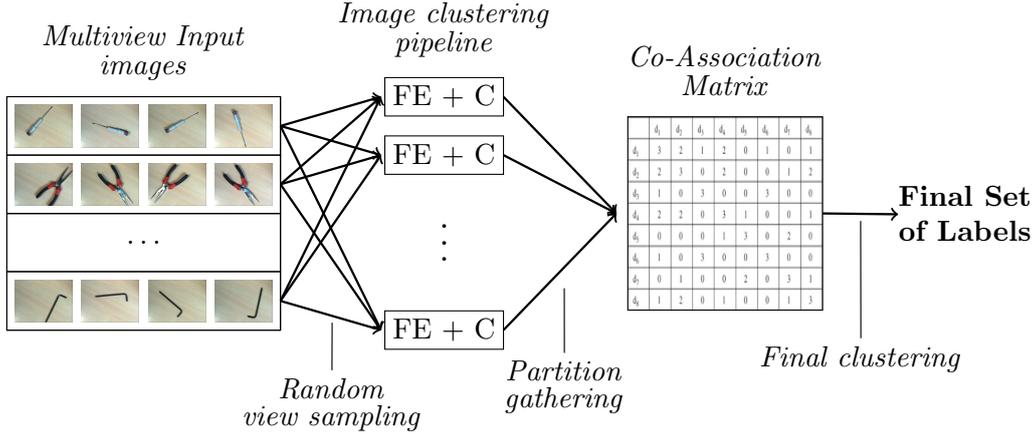

%% file: table_multi_robustness.tex
\begin{table}[!ht]
\centering
\caption{Clustering results of MVEC for different BLC. For comparison, we remind the corresponding results using single views, in parenthesis.}
\label{tab:multiview}

\vspace{\baselineskip}
    \begin{tabular}{c|c"c|c}
        
        \multicolumn{2}{c"}{} & NMI & Purity \tabularnewline
        \Xhline{2\arrayrulewidth}
        \multicolumn{2}{c"}{BLC1} & 0.95 \footnotesize{\textit{(0.86)}} & 0.96 \footnotesize{\textit{(0.85)}} \tabularnewline
        \multirow{5}{*}{BLC2} & Very dark & 0.91 \footnotesize{\textit{(0.77)}} & 0.93 \footnotesize{\textit{(0.77)}} \tabularnewline
        & Dark & 1.00 \footnotesize{\textit{(0.88)}} & 1.00 \footnotesize{\textit{(0.89)}} \tabularnewline
        & Normal & 1.00 \footnotesize{\textit{(0.90)}} & 1.00 \footnotesize{\textit{(0.90)}} \tabularnewline
        & Bright & 0.96 \footnotesize{\textit{(0.84)}} & 0.96 \footnotesize{\textit{(0.84)}} \tabularnewline
        & Very bright & 0.84 \footnotesize{\textit{(0.73)}} & 0.86 \footnotesize{\textit{(0.74)}} \tabularnewline
        \multicolumn{2}{c"}{BLC3} & 0.95 \footnotesize{\textit{(0.84)}} & 0.96 \footnotesize{\textit{(0.85)}} \tabularnewline
        \multicolumn{2}{c"}{BLC4} & 0.84 \footnotesize{\textit{(0.69)}} & 0.82 \footnotesize{\textit{(0.69)}} \tabularnewline
        \multicolumn{2}{c"}{BLC5} & 0.95 \footnotesize{\textit{(0.83)}} & 0.96 \footnotesize{\textit{(0.81)}} \tabularnewline
    \end{tabular}
    
\end{table}

%% file: conclusion.tex
\section{Conclusion}
\label{sec:conclusion}

\subsection{Conclusive remarks} 
In this paper, we introduced a new kind of robotics pick-and-place application, called Unsupervised Robotic Sorting, which consists in physically grouping together previously unseen objects in a way that makes sense at a human level. We propose an implementation in which objects are represented by images and the unsupervised classification module is composed of a pretrained Xception feature extractor and Agglomerative Clustering. This pipeline was shown to be good through experiments on several public datasets, as well as a specifically created robustness testing dataset. This approach is shown to work in a practical implementation, in a real-world industrial environment.

The robustness testing dataset, which is challenging for image-set clustering, is made publicly available to help other researchers to test their image clustering algorithms. We also show, through further experimentation on this dataset, that using multiple images for each object can increase URS performance. By embedding the chosen image clustering pipeline into a co-association ensemble clustering framework, we obtain more accurate and more robust clustering results.

\subsection{Future working directions} 
The proposed approach to solve URS demonstrates good results and is very encouraging. However, we have also shown that URS would benefit from using multiple views. Future work will consist in implementing such multi-view URS on a real robot. To do so, we first need to solve another problem of view selection. We have already started to investigate the question of autonomous optimal view selection \cite{iros}, by training a neural network to predict the camera pose which is best suited for clustering. Further developments will consist in combining this approach with the multi-view framework proposed in this paper.

In this paper, the focus is placed on the decision making module of the autonomous sorting application. For this reason, the scene segmentation, data gathering and grasping were simple fixed modules. Now that the intelligent unsupervised sorting module has been shown to work, it will be interesting to include it in a more complete pick-and-place pipeline, which handles automatic scene segmentation and robotic grasp detection.